%% file: jid.tex
\documentclass[10pt,onecolumn]{article}

\usepackage{times}
\usepackage{epsfig}
\usepackage{graphicx}
\usepackage{amsmath}
\usepackage{amssymb,color,bm}
\usepackage[utf8]{inputenc} 
\usepackage{caption}
\usepackage{subcaption}
\usepackage{booktabs}
\usepackage{multirow}
\usepackage{epstopdf}
\usepackage{mathtools}
\usepackage{todonotes}
\usepackage{tikz,pgfplots}
\usepackage{xspace}
\usepackage[pagebackref=true,breaklinks=true,colorlinks,bookmarks=false,urlcolor=blue]{hyperref}
\usepackage{geometry}

\def\abstract
   {%
   \centerline{\large\bf Abstract}%
   \vspace*{12pt}%
   \it%
   }

\def\affiliation#1{\gdef\@affiliation{#1}} \gdef\@affiliation{}

\makeatletter
\DeclareRobustCommand\onedot{\futurelet\@let@token\@onedot}
\def\@onedot{\ifx\@let@token.\else.\null\fi\xspace}

\def\etal{\emph{et al}\onedot}
\makeatother

\newcommand{\AO}{\boldsymbol{\Omega}}
\newcommand{\OB}{\mathrm{OB}}

\begin{document}

\title{A Joint Intensity and Depth Co-Sparse Analysis Model \\for Depth Map Super-Resolution}
\date{}
\author{Martin Kiechle, Simon Hawe, and Martin Kleinsteuber\\
Department of Electrical Engineering and Information Technology,\\
Technische Universität M\"unchen, Munich, Germany\\
{\tt\small \{martin.kiechle,simon.hawe,kleinsteuber\}@tum.de} \\
{\small\url{http://www.gol.ei.tum.de}}
}

\maketitle

\begin{abstract}
High-resolution depth maps can be inferred from low-resolution depth measurements and an additional high-resolution intensity image of the same scene. To that end, we introduce a bimodal co-sparse analysis model, which is able to capture the interdependency of registered intensity and depth information. This model is based on the assumption that the co-supports of corresponding bimodal image structures are aligned when computed by a suitable pair of analysis operators. 
No analytic form of such operators exist and we propose a method for learning them from a set of registered training signals.
This learning process is done offline and returns a bimodal analysis operator that is universally applicable to natural scenes.
We use this to exploit the bimodal co-sparse analysis model as a prior for solving inverse problems, which leads to an efficient algorithm for depth map super-resolution.
\end{abstract}

\section{Introduction}
\label{sec:introduction}
\input{sections/10_introduction.tex}

\section{Related Work}
\label{sec:related_work}
\input{sections/20_previouswork.tex}

\section{Proposed Approach}
\label{sec:approach}
\input{sections/30_analysisoperators.tex}

\section{Results and Comparison}
\label{sec:results}
\input{sections/40_results.tex}

\section{Conclusion and Discussion}
\label{sec:conclusion}
\input{sections/50_discussion.tex}

{\small
\bibliographystyle{ieee}
\bibliography{PaperCitations}
}

\end{document}

%% file: sections/10_introduction.tex
Many technical applications in fields like robotics, 3D video rendering, or human computer interaction are built upon precise knowledge of the surrounding 3D environment. This information is typically acquired either via passive or active range sensors.
%
%
Passive range sensing, i.e.\ 3D from stereo intensity images, is essentially based on three steps. First, ambient light that is reflected from the same object surfaces is captured at multiple displaced views. Second, the disparities of corresponding light intensity samples between the different views are determined. Third, the distance to the sensor is obtained using the computed disparities together with the knowledge of the relative positions between all views. Despite very active research in this area and significant improvements over the past years, stereo methods still struggle with noise, texture-less regions, repetitive texture, and occluded areas. For an overview of stereo methods, the reader is referred to \cite{Seitz}.

\begin{figure}[t]
	\centering
	\begin{subfigure}[b]{0.488\columnwidth}
		\centering
		\includegraphics[width=\textwidth]{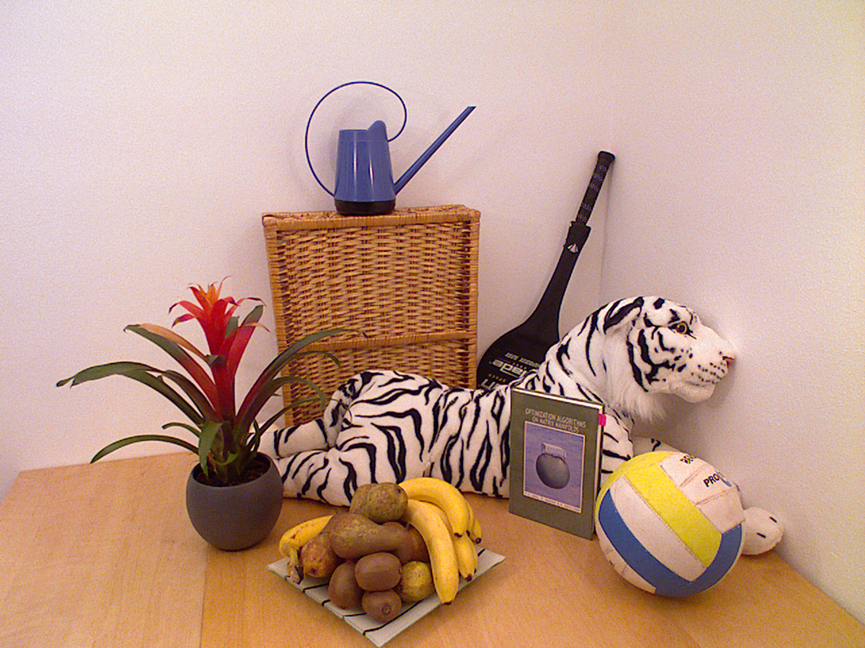}
	\end{subfigure}%
	\,
	\begin{subfigure}[b]{0.488\columnwidth}
		\centering
		\includegraphics[width=\textwidth]{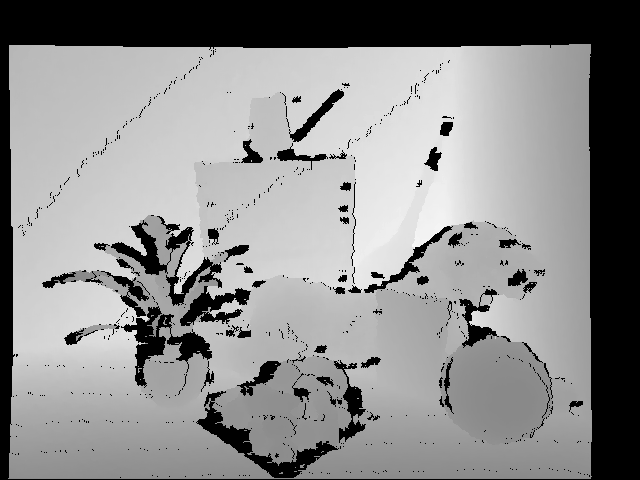}
	\end{subfigure}%
	
	\medskip
	\begin{subfigure}[b]{0.488\columnwidth}
		\centering
		\includegraphics[width=\textwidth]{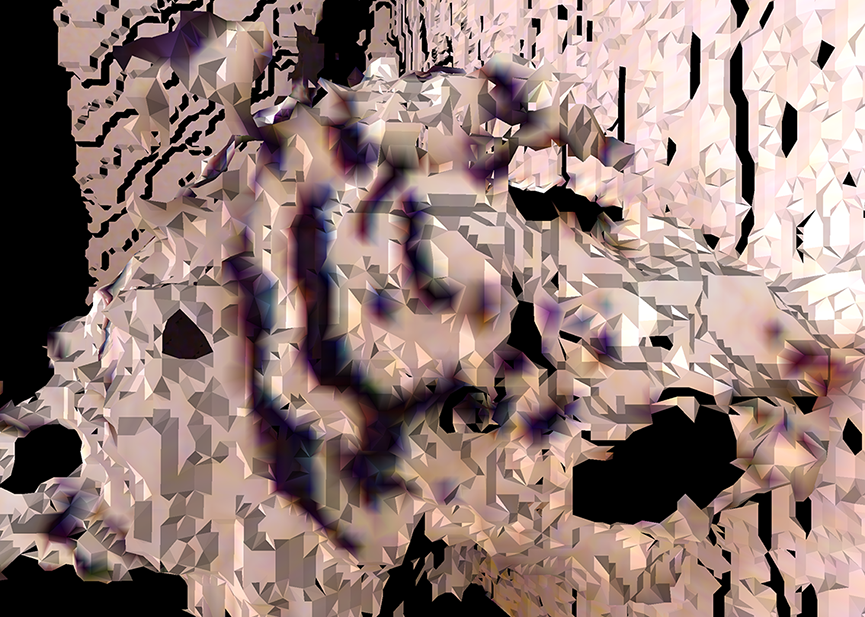}
	\end{subfigure}%
	\,
	\begin{subfigure}[b]{0.488\columnwidth}
		\centering
		\includegraphics[width=\textwidth]{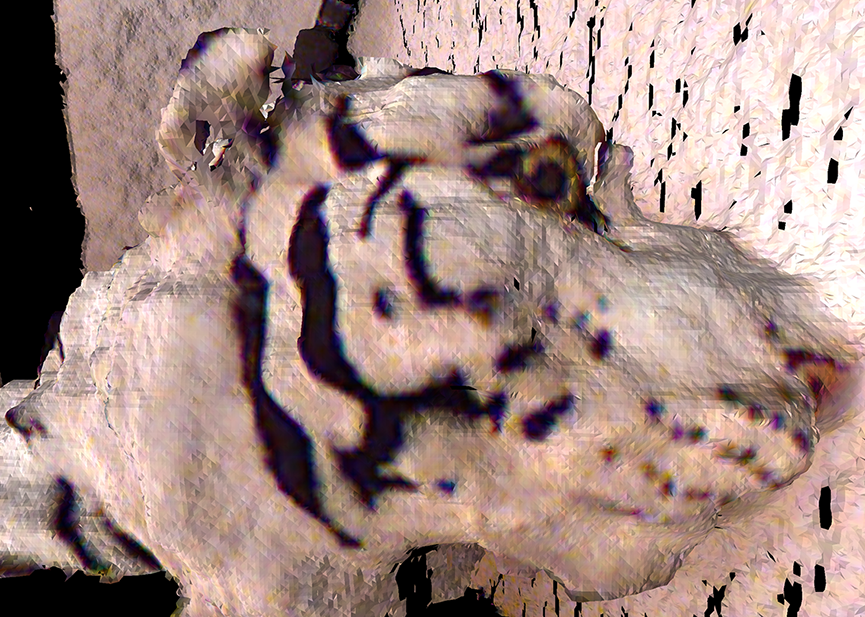}
	\end{subfigure}%
	\caption{Top row: color image (left) and corresponding registered depth map (right) recorded by the Kinect sensor. Bottom row: a 3D rendering of the tiger head detail visualizing the difference between the original sensor data (left) and the refined version using our proposed method (right).}
	\label{fig:eye_catcher}
\end{figure}

Active sensors, on the other hand, emit light and either measure the time-of-flight of a modulated ray, e.g.\ LIDAR or PMD, or capture the reflection pattern of a structured light source to infer the distance to objects, as is done for example by the well-known Microsoft Kinect. Such sensors become more and more popular, because they acquire reliable depth measurements independent of the occurring texture and are real-time capable. However, the main drawbacks are that the acquired depth maps are of low-resolution (LR) and corrupted by noisy and missing values. To overcome these limitations, different methods for upsampling and denoising LR depth maps from range sensors have been proposed, see Section \ref{sec:related_work}. 


A co-occurrence of signal patterns in both the depth map obtained by an active range sensor as well as in a corresponding registered camera intensity image, is suggested by the fact that both ambient and artificially emitted light is reflected by the same object surfaces. Indeed, some of the most successful methods for reconstructing and refining depth maps aim at exploiting this statistical dependency. 

In this paper, we introduce a joint intensity and depth (JID) co-sparse analysis model that exploits the dependencies between the two modalities. This model is based on the assumption that the co-supports of corresponding structures are aligned when computed by a suitable pair of analysis operators. To that end, we propose a method for learning the required bimodal analysis operator from aligned training data.
This procedure is done only once and offline, and results in a universally applicable operator, which is valid for all intensity and depth pairs of natural scenes. This operator together with a high-resolution (HR) intensity image is employed for reconstructing a HR depth map that corresponds to the HR intensity image. The problem is considered as a linear inverse problem, which is regularized using the bimodal analysis operator. Our numerical experiments show that our method compares favorably to state-of-the-art methods both visually and quantitatively, and they underpin the validity of our proposed joint intensity and depth data model.
In summary, the two main contributions of this paper are:
\begin{itemize}
\item The new bimodal co-sparse analysis model that reflects the dependencies between properly aligned intensity and depth samples from the same scene. 
\item An algorithm for simultaneous depth map super-resolution (SR) and inpainting of missing depth values, which exploits the introduced data model and allows to cope with various noise models.
\end{itemize}

%
%


%% file: sections/20_previouswork.tex
Increasing the resolution of depth images obtained from range sensors has become an important research topic, and diverse approaches treating this problem have been proposed throughout the past years. Many of these methods originate from the closely related problem of intensity image super-resolution. However, these mostly aim at producing pleasantly looking results, which is different from the goal of achieving geometrically sound depth maps. Straightforward upsamling methods like nearest-neighbor, bilinear, or bicubic interpolation produce undesirable staircasing or blurring artifacts, see Figure \ref{fig:tsukuba_comparison}. Here, we shortly review more sophisticated methods for depth map SR that aim at reducing these artifacts.
 
In a first attempt, methods have been proposed that use smoothing priors from edge statistics \cite{Fattal2007} or local self-similarities \cite{Freedman2011}. These methods only require a single image, but either have difficulties in textured areas, or only work well for small upscaling factors.
%
%
A different approach, which also solely requires depth information is based on fusing multiple displaced LR depth maps into a single HR depth map. Schuon \etal \cite{Schuon2009} develop a global energy optimization framework employing data fidelity and geometry priors. This idea is extended for better edge-preservation by Bhavsar \etal in \cite{Bhavsar2012}.

A number of recently introduced methods aim at exploiting co-aligned discontinuities in intensity and depth images of the same scene. They fuse the HR and LR data utilizing Markov Random Fields (MRF). Depth map refinement based on MRF has been first explored in \cite{Diebel2006}, extended in \cite{Lu2011a} with a depth specific data term, and combined with depth from passive stereo in \cite{Zhu2008}. In order to better preserve local structures and to remove outliers, Park \etal \cite{Park2011} add a non-local means term to their MRF formulation. Aodha \etal \cite{Aodha2012} treat depth SR as an MRF labeling problem of matching LR depth map patches to HR patches from a predefined database.

Inspired by successful stereo matching algorithms, Yang \etal \cite{Yang2007} iteratively employ a bilateral filter to improve depth SR using an additional HR intensity image. Chan \etal \cite{Chan2008} extend this approach by incorporating a noise model specific to depth data. Xiang \etal \cite{Xiang2010} include sub-pixel accuracy, and Dolson \etal \cite{Dolson2010} address temporal coherence across a depth data stream from LIDAR scanners by combining a bilateral filter with a Gaussian framework.

Finally, methods exist that exploit the dependency between sparse representations of intensity and depth signals over appropriate dictionaries. In \cite{Gudmundsson2011}, the complex wavelet transform is used as the dictionary. Both the HR intensity image and the LR depth map are transformed into this domain and the resulting 
coefficients  
are fused using a dual tree to obtain the HR depth map. Instead of using predefined bases, approaches employing learned dictionaries are known to lead to state-of-the-art performance in diverse classical image reconstruction tasks, cf. \cite{Elad2010,Mairal2008}. Surprisingly, applying those techniques for depth map enhancement has only very recently been explored. Mahmoudi \etal \cite{Mahmoudi2012a} first learn a depth dictionary from noisy samples, then refine and denoise these samples and finally learn an additional dictionary from the denoised samples to inpaint, denoise, and super-resolve projected depth maps from 3D models. Closest to our approach are the recent efforts of \cite{Li2012} and \cite{Tosic2012}. They independently learn dictionaries of depth and intensity samples, and model a coupling of the two signal types during the reconstruction phase. In \cite{Li2012}, three dictionaries are composed from LR depth, HR depth, and HR color samples to learn a respective mapping function based on edge features. In contrast, only two dictionaries for intensity and depth are learned in \cite{Tosic2012}, where the similarity of the support of corresponding sparse representations is used to model the coupling.   
%

%% file: sections/30_analysisoperators.tex
In our approach, we treat the problem of depth map super-resolution as a linear inverse problem. Basically, the goal is to reconstruct a HR depth map $\mathbf{s}\in\mathbb{R}^{n}$ from a set of measurements $\mathbf{y}\in\mathbb{R}^{m}$ that are possibly corrupted by noise and missing values, i.e. a LR depth map, with $m \leq n$. Formally, the relation between $\mathbf{s}$ and $\mathbf{y}$ is given by
%
\begin{equation}\label{eq:inv_problem}
\mathbf{y} = \mathcal{A}\mathbf{s}+\mathbf{e},
\end{equation}
with $\mathcal{A}\in\mathbb{R}^{m \times n}$ modeling the sampling process, and $\mathbf{e}\in\mathbb{R}^{m}$ modeling noise and potential sampling errors. Here, the dimension $m$ of the measurement vector is significantly smaller than the dimension $n$ of the HR depth map. Consequently, reconstructing $\mathbf{s}$ in \eqref{eq:inv_problem} is highly ill-posed. Using additional information about the signal's structure helps to tackle this linear inverse problem.

One prior assumption that has proven useful, is that the signals of interest allow a sparse representation. A vector is called sparse, when most of its entries are equal to zero or sufficiently small in magnitude. The \emph{co-sparse analysis model} \cite{nam:2011} assumes that applying an analysis operator $\AO \in \mathbb{R}^{k\times n}$ with $k \geq n$ to a signal $\mathbf{s} \in \mathbb{R}^n$ results in a sparse vector $\AO  \mathbf{s} \in \mathbb{R}^k$. If $g\colon \mathbb{R}^{k} \to \mathbb{R}$ denotes a function that measures sparsity like the $\ell_0$-pseudo-norm, the analysis model assumption can be exploited to tackle linear inverse problems by solving
\begin{equation}\label{eq:analysis_model}
\mathbf{s}^{\star} \in \arg\underset{\mathbf{s}\in \mathbb{R}^{n}}{\min} \,g(\mathbf{\Omega s})\; \text{subject to}\; d_{E}(\mathcal{A}\mathbf{s},\mathbf{y}) \leq \varepsilon,  
\end{equation}
where $d_{E}$ denotes an appropriate error measure and $\varepsilon \in \mathbb{R}^{+}_0$ is an estimated upper bound of the noise energy. Typical examples for $d_{E}$ include the squared Euclidean distance.

Most crucial for the success of the analysis approach is the choice of an appropriate analysis operator. Analytic operators, e.g.\ the finite difference operator, exist. However, using an operator that is learned from signal examples is known to yield better performance \cite{Hawe2012, Ophir2011, Rubinstein2012, Yaghoobi2011}.

Our approach to depth map SR utilizes the interdependency of the two modalities intensity and depth. In a first step we describe a new data model and how it can be learned in the form of an analysis operator pair that incorporates both, signal structure and their according bimodal interdependency. In a second step, we explain how this learned prior model can be used for HR signal reconstruction.

\subsection{Bimodal Co-Sparse Analysis Model}
\label{sec:cosparse_analysis_model}
\input{sections/31_cosparsemodel.tex}

\subsection{JID Analysis Operator Learning}
\label{sec:operator_learning}
\input{sections/32_operatorlearning.tex}
\subsection{Depth Map Super-Resolution}
\label{sec:depth_SR}
\input{sections/33_reconstruction.tex}

%% file: sections/31_cosparsemodel.tex
%
In the analysis model, the zero entries of the analyzed vector $\AO \mathbf{s}$ determine the signal's structure \cite{nam:2011}. Geometrically, $\mathbf{s}$ lies in the intersection of all hyperplanes whose normal vectors are given by the rows of $\AO$ indexed by the zero entries of $\AO \mathbf{s}$. This index set is called the \emph{co-support} of $\mathbf{s}$, and is given by 
\begin{equation}\label{eq:cosupp}
cosupp(\AO \mathbf{s}):=\{j~|~(\AO \mathbf{s})_j=0\}.
\end{equation}
Therein, $\mathbf{s}$ is a vectorized patch and $(\AO\mathbf{s})_j$ is the $j$-th entry of the analyzed vector.
%
Now assume that intensity signals $\mathbf{s}_I \in \mathbb{R}^{n_1}$ as well as depth signals $\mathbf{s}_D \in \mathbb{R}^{n_2}$ allow a co-sparse representation with an appropriate pair of analysis operators $(\AO_I, \AO_D)\in \mathbb R^{k \times n_1}\times \mathbb R^{k \times n_2}$. 
Based on the knowledge that a signal's structure is encoded in its co-support \eqref{eq:cosupp}, we postulate that 
\emph{a pair of analysis operators exists such that the co-support of $\mathbf{s}_I$ and $\mathbf{s}_D$ are statistically dependent, if both signals originate from the same scene}. The bimodal co-sparse analysis model assumes that the conditional probability of $j$ belonging to the co-support of $\mathbf{s}_D$ given that $j$ belongs to the co-support of $\mathbf{s}_I$ is significantly higher than the unconditional probability, i.e.
\begin{equation}\label{eq:cond_prob}
\begin{split}
Pr( \{j \in cosupp(\AO_D \mathbf{s}_D) \}~|~\{j \in cosupp(\AO_I \mathbf{s}_I) \}) \gg Pr(\{j \in cosupp(\AO_D \mathbf{s}_D) \}).
\end{split}
\end{equation}
Clearly, this model is idealized, since in practice, the entries of the analyzed vectors are not exactly equal to zero. In the next section, we explain how the coupled pair of analysis operators $(\AO_I, \AO_D)$ can be jointly learned, such that aligned intensity and depth signals analyzed by these operators adhere to the introduced model.

%% file: sections/32_operatorlearning.tex
%
%
%
Generally, the goal of learning an analysis operator can be formulated as follows: Given a set ${\left\lbrace \mathbf{s}^{(i)} \in \mathbb{R}^n \right\rbrace}_{i=1}^M$ of training samples representing the signal class of interest, find an operator $\AO \in \mathbb{R}^{k \times n}$ with $k \geq n$ such that all representations $\AO \mathbf{s}^{(i)}$ are maximally sparse. 

%
%
Here, we aim at learning the coupled pair of bimodal analysis operators $(\AO_I, \AO_D)\in \mathbb{R}^{k\times n_1}\times \mathbb{R}^{k \times n_2}$ for intensity and depth signals. Therefore,
we use a set of $M$ aligned  and corresponding training pairs $\{(\mathbf{s}_I^{(i)},\mathbf{s}_D^{(i)})\in\mathbb{R}^{n_1} \times \mathbb{R}^{n_2}\}_{i=1}^M$.
More specifically, these are HR intensity and HR depth patches representing the same excerpt of a scene.
Now, we incorporate the proposed condition \eqref{eq:cond_prob} into the learning process by enforcing the zeros of corresponding analyzed vectors $\AO_I \mathbf{s}_{I}^{(i)},\AO_D \mathbf{s}_{D}^{(i)}$ to be at the same positions. Throughout the paper, the function
$\mathbf{x} \mapsto \sum_{j=1}^k \log(1 + \nu x_j^2),$
with $\nu >0$ being a positive weight, serves as an appropriate sparsity measure. Note, that any other smooth sparsity measure principally leads to similar results. With this, the coupled sparsity is controlled through the function
\begin{equation}\label{eq:coupling}
g(\AO_I \mathbf{s}_I^{(i)},\AO_D\mathbf{s}_D^{(i)}) := \sum \limits_{j=1}^{k} \log \bigg(1 + \nu\big((\AO_I \mathbf{s}_{I}^{(i)})_j^2 + (\AO_D \mathbf{s}_{D}^{(i)})_j^2\big)\bigg).
\end{equation}
%
To find the ideal pair of bimodal operators we minimize the sum of squares of \eqref{eq:coupling}, which can be interpreted as a balanced optimization over the expectation and the variance of the analyzed vectors' sparsity and reads as
\begin{equation}\label{eq:coupling_mat}
G(\AO_I,\AO_D)  := 
\tfrac{1}{M} \sum \limits_{i=1}^{M} g(\AO_I \mathbf{s}_I^{(i)},\AO_D\mathbf{s}_D^{(i)})^2.
\end{equation}
Additionally, we take separate constraints on the operator into account which are motivated in \cite{Hawe2012} and summarized in the following.
%

The possible solutions of the transposed of a single analysis operator are restricted to the set of full-rank matrices with normalized columns, known as the oblique manifold $\OB(n,k)$.
Since $\OB(n,k)$ is open and dense in the set of matrices with normalized columns, the penalty function
\begin{equation}\label{eq:logdet}
h(\AO):= -\tfrac{1}{n\log(n)} \log \det(\tfrac{1}{k} \AO^\top \AO)
\end{equation}
is used to adhere to the rank condition and to prevent iterates to approach the boundary of $\OB(n,k)$. 
Furthermore, a penalty function is incorporated that enforces the operators to have distinctive rows, and which controls the mutual coherence of each operator
\begin{equation}\label{eq:lindep}
r(\AO):= - \hspace{-4mm} \sum \limits_{1 \leq i < l \leq k} \log(1-({\bm \omega}_{i}^\top {\bm \omega}_{l})^2),
\end{equation}
with ${\bm \omega}_{i}$ denoting the transposed of the $i$-th row of $\AO$.

Combining the two penalties into 
$p(\AO):= \kappa h(\AO) + \mu r(\AO),$
with $\kappa,\mu \in \mathbb{R}^+$ being positive weights, and using $n_1=n_2=:n$ for legibility reasons, 
our problem of learning the pair of JID analysis operators is given by
\begin{equation}
\begin{split}
\left( \AO_{I}^{\top}, \AO_{D}^{\top} \right) \in \underset{\mathcal{X}_I ,\mathcal{X}_{D} \in \OB(n,k)}{\arg\min} & G(\mathcal{X}^{\top}_{I},\mathcal{X}^{\top}_{D}) + p(\mathcal{X}^{\top}_{I}) + p(\mathcal{X}^{\top}_{D}).
\end{split}
\label{eq:bimodal_OP}
\end{equation}
The arising optimization problem is solved with a geometric CG method using an Armijo step size rule, cf.\ \cite{abs:oamx08}.

For the evaluation of our approach we train one fixed operator pair and use it in all presented experiments. To that end, we
gather a total of $M=15000$ pairs of squared sample patches of size $\sqrt{n}=5$ from the five registered intensity and depth image pairs 'Baby1', 'Bowling1', 'Moebius', 'Reindeer' and 'Sawtooth' of the Middlebury stereo set. As it is common in dictionary learning methods, we require all training patches to have zero-mean.
Furthermore, we learn the operators with twofold redundancy, i.e.\ $k=2n$, resulting in the operator pair $(\AO_{I}, \AO_{D}) \in \mathbb{R}^{50 \times 25} \times \mathbb{R}^{50 \times 25}$. In general, a larger redundancy of the operators leads to better reconstruction quality but at the cost of increased computational complexity of both learning and reconstruction. Twofold redundancy provides a good trade-off between reconstruction quality and computation time. We empirically set the remaining parameters to $\nu=10$, $\kappa=9\cdot 10^4$ and $\mu=10^2$.

%% file: sections/33_reconstruction.tex
In this section, we explain how the pair of patch based bimodal analysis operators $(\AO_I, \AO_D)$
is used to jointly reconstruct an aligned pair of intensity and depth signals $\mathbf{s}_I,\mathbf{s}_D \in \mathbb{R}^N$ from a set of measurements  $\mathbf{y}_I \in \mathbb{R}^{m_1},\mathbf{y}_D \in \mathbb{R}^{m_2}$. Here $\mathbf{s}_I,\mathbf{s}_D$ are the vectorized versions of an HR intensity image and an HR depth map obtained by ordering their entries lexicographically, with $N=wh$ where $w$ and $h$ denote the height and width of both HR signals.

To use our bimodal operator for reconstructing entire images or depth maps, we need to extend the application of the operator beyond local patches. To achieve this, we recall the approach in \cite{Hawe2012} for the unimodal case.
Instead of reconstructing each patch individually and combining them in a final step to form the image, the complete $N$-dimensional signal is reconstructed by minimizing the average sparsity of all patches. In this way, neighboring patches support each other during the optimization process.
Accordingly, a \emph{global} analysis operator $\AO^{F} \in \mathbb{R}^{K \times N}$ is constructed from a patch based operator $\AO \in \mathbb{R}^{k \times n}$. Therefore, let $\mathcal{P}_{rc} \in \mathbb{R}^{n \times N}$ denote the operator, which selects the ($\sqrt{n} \times \sqrt{n}$)-dimensional patch centered at position $(r,c)$ from the signal, then the global operator is given as
\begin{equation} \label{eq:op}
\AO^{F} := 
{\small
\begin{bmatrix} 
\AO\mathcal{P}_{11} \\ 
\AO\mathcal{P}_{21} \\ 
\vdots 																							\\ 
\AO\mathcal{P}_{hw} 
\end{bmatrix}}
\in \mathbb{R}^{K \times N},
\end{equation}
with $K = whk$, i.e.\ all patch positions are considered. The reflective boundary condition is used to deal with problems along boundaries.

Now, with the global operator pair $(\AO_I^F,\AO_D^F)$, the bimodal extension of the signal reconstruction in \eqref{eq:analysis_model} is given by
\begin{equation}
\begin{split}
& (\mathbf{s}^\star_I,\mathbf{s}^\star_D) \in \arg \underset{\mathbf{s}_I,\mathbf{s}_D\in \mathbb{R}^{N}} {\min} g(\AO^{F}_I\mathbf{s}_I,\AO^{F}_D\mathbf{s}_D) \label{eq:image_reconstruction_basic}\\
& \text{subject to} \quad d_{E} \left( \left( \mathcal{A}_I\mathbf{s}_I, \mathcal{A}_D\mathbf{s}_D \right), \left( \mathbf{y}_I, \mathbf{y}_D \right) \right)  \leq \varepsilon. 
\end{split}
\end{equation}
Therein, the sparsity measure $g$ is the same as the one in Equation \eqref{eq:coupling}. Consequently, the analyzed versions of both modalities are enforced to have a correlated co-support and hence the two signals are coupled. 

The measurement matrices $\mathcal{A}_I \in \mathbb{R}^{m_1 \times N}$ and $\mathcal{A}_D \in \mathbb{R}^{m_2 \times N}$ model the sampling process of each modality.
Here, we focus on enhancing the quality of depth measurements $\mathbf{y}_D$, given a fixed high quality intensity signal $\mathbf{y}_I=\mathbf{s}_I$ by simultaneously upsampling and inpainting missing measurements. In this case, $\mathcal{A}_I$ is the identity operator 
and the analyzed intensity signal is constant, i.e. $\AO^{F}_I\mathbf{s}_I=\mathbf{c}=\textit{const.}$ This simplifies Problem \eqref{eq:image_reconstruction_basic} for recovering a HR depth map to
\begin{equation}
\begin{split}
& \mathbf{s}^\star_D \in \arg\underset{\mathbf{s}_D\in \mathbb{R}^{N}} {\min} \, g(\mathbf{c},\AO^{F}_D\mathbf{s}_D) \label{eq:depth_reconstruction}\\
& \text{subject to} \quad d_{E}(\mathcal{A}_D\mathbf{s}_D,\mathbf{y}_D) \leq \varepsilon_{D}.
\end{split}
\end{equation}
The data fidelity term $d_{E}$ depends on the error model of the depth data and can be chosen accordingly. For instance, this may be an error measure tailored to a sensor specific model, cf.\ Section \ref{sec:kinect_validation}.
%
In this way, knowledge about the scene gained from the intensity image and its co-support regarding the bimodal analysis operators helps to determine the HR depth signal.
%
%


%% file: sections/40_results.tex
In this section we experimentally evaluate our approach by conducting two sets of experiments. First, we evaluate our approach numerically on synthetic data using the well-known Middlebury stereo dataset \cite{Scharsteina}, which provides aligned intensity images and depth maps for a number of different test scenes. Second, we evaluate our method on real-world data by processing scenes captured with the popular Microsoft Kinect sensor.

\subsection{Quantitative Evaluation}
\label{sec:middlebury_evaluation}
\input{sections/41_middlebury.tex}

\subsection{Validation on Kinect Data}
\label{sec:kinect_validation}
\input{sections/42_kinect.tex}

%% file: sections/41_middlebury.tex
To compare our results to the state-of-the-art, we quantitatively evaluate our algorithm on the four standard test images 'Tsukuba', 'Venus', 'Teddy', and 'Cones' from the Middlebury dataset. To artificially create LR input depth maps, we scale the ground truth depth maps down by a factor of $d$ in both vertical and horizontal dimension. We first blur the available HR image with a Gaussian kernel of size $(2d-1)\times(2d-1)$ and standard deviation $\sigma=d/3$ before downsampling. The LR depth map and the corresponding HR intensity image are the input to our algorithm.

\begin{table}[t]
  \centering
  \resizebox{0.6\linewidth}{!}{
	  \begin{tabular}{rrcccc}
	    \toprule
	        $d$ & method & Tsukuba & Venus & Teddy & Cones \\
	    \midrule
	    \multicolumn{1}{c}{\multirow{4}[0]{*}{2x}} & nearest-neighbor & 1.24 & 0.37	& 4.97 & 2.51	\\
	    \multicolumn{1}{c}{} & Yang \etal \cite{Yang2007} & 1.16  & 0.25  & 2.43  & 2.39 \\        
	    \multicolumn{1}{c}{} & Hawe \etal \cite{Hawe2012} & 1.03 & 0.22 & 2.95 & 3.56 \\
	    \multicolumn{1}{c}{} & our method & \textbf{0.47} & \textbf{0.09} & \textbf{1.41} & \textbf{1.81} \\
	    \midrule
	    \multicolumn{1}{c}{\multirow{4}[0]{*}{4x}} & nearest-neighbor & 3.53 & 0.81	& 6.71 & 5.44	\\ 
	    \multicolumn{1}{c}{} & Yang \etal & 2.56  & 0.42  & 5.95  & \textbf{4.76} \\
	    \multicolumn{1}{c}{} & Hawe \etal & 2.95  & 0.65  & 4.80  & 6.54 \\
	    \multicolumn{1}{c}{} & our method & \textbf{1.73} & \textbf{0.25} & \textbf{3.54} & 5.16 \\
	    \midrule
	    \multicolumn{1}{c}{\multirow{5}[0]{*}{8x}} & nearest-neighbor & 3.56 & 1.90 & 10.9 & 10.4 \\
	    \multicolumn{1}{c}{} & Yang \etal & 6.95  & 1.19  & 11.50 & 11.00 \\   
	    \multicolumn{1}{c}{} & Lu \etal \cite{Lu2011a} & 5.09  & 1.00  & 9.87  & 11.30 \\
	    \multicolumn{1}{c}{} & Hawe \etal & 5.59  & 1.24  & 11.40 & 12.30 \\
	    \multicolumn{1}{c}{} & our method & \textbf{3.53} & \textbf{0.33} & \textbf{6.49} & \textbf{9.22} \\
	    \bottomrule
	  \end{tabular}%
  }
  \caption{Numerical comparison of our method to other depth map SR approaches for different upscaling factors $d$. The figures represent the percentage of bad pixels with respect to all pixels of the ground truth data and an error threshold of $\delta=1$.}
  \label{tab:middlebury_results}%
\end{table}%

\begin{table}[t]
  \centering
  \resizebox{0.6\linewidth}{!}{
	  \begin{tabular}{rrcccc}
	    \toprule
				$d$ & method & Tsukuba & Venus & Teddy & Cones \\
	    \midrule
	    \multicolumn{1}{c}{\multirow{5}[0]{*}{2x}} & nearest-neighbor & 0.612 & 0.288 & 1.543 & 1.531 \\
			\multicolumn{1}{c}{} & Chan \etal \cite{Chan2008} & n/a & 0.216 & 1.023 & 1.353 \\   
	    \multicolumn{1}{c}{} & Aodha \etal \cite{Aodha2012}& 0.601 & 0.296 & 0.977  & 1.227 \\
	    \multicolumn{1}{c}{} & Hawe \etal \cite{Hawe2012}& 0.278 & 0.105 & 0.996  & 0.939 \\
	    \multicolumn{1}{c}{} & our method & \textbf{0.255} & \textbf{0.075} & \textbf{0.702} & \textbf{0.680}  \\
	    \midrule
	    \multicolumn{1}{c}{\multirow{5}[0]{*}{4x}} & nearest-neighbor & 1.189 & 0.408 & 1.943 & 2.470 \\
			\multicolumn{1}{c}{} & Chan \etal & n/a & 0.273 & \textbf{1.125} & 1.450 \\
	    \multicolumn{1}{c}{} & Aodha \etal & 0.833 & 0.395 & 1.184 & 1.779 \\
	    \multicolumn{1}{c}{} & Hawe \etal & \textbf{0.450} & 0.179 & 1.389 & 1.398 \\
	    \multicolumn{1}{c}{} & our method & 0.487 & \textbf{0.129} & 1.347 & \textbf{1.383} \\
	    \midrule
	    \multicolumn{1}{c}{\multirow{4}[0]{*}{8x}} & nearest-neighbor & 1.135 & 0.546 & 2.614 & 3.260 \\
			\multicolumn{1}{c}{} & Chan \etal & n/a & 0.369 & \textbf{1.410} & \textbf{1.635} \\
	    \multicolumn{1}{c}{} & Hawe \etal & \textbf{0.713} & 0.249 & 1.743 & 1.883 \\
	    \multicolumn{1}{c}{} & our method & 0.753 & \textbf{0.156} & 1.662 & 1.871 \\
	    \bottomrule
	  \end{tabular}%
  }
\caption{Numerical comparison of our method to other depth map SR approaches. The figures represent the RMSE in comparison with the ground truth depth map.}
\label{tab:rmse_results}%
\end{table}%

Here, we assume an i.i.d.\ normal distribution of the error, which leads to
the data fidelity term \linebreak $d_{E}(\mathcal{A}_D\mathbf{s}_D,\mathbf{y}_D)=\Vert\mathcal{A}_D\mathbf{s}_D-\mathbf{y}_D\Vert ^2_2$. From \eqref{eq:depth_reconstruction} we get the unconstrained optimization problem for reconstructing the HR depth signal as
\begin{equation}
\mathbf{s}^\star_D \in \underset{\mathbf{s}_D\in \mathbb{R}^{N}} {\arg\min} \, \lambda g(\mathbf{c},\AO^{F}_D\mathbf{s}_D) + \Vert\mathcal{A}_D\mathbf{s}_D-\mathbf{y}_D\Vert ^2_2.
\label{eq:depth_reconstruction_middlebury}
\end{equation}
Larger values of the weighting factor $\lambda \propto \varepsilon^{-1}_D$ lead to a faster convergence of the algorithm but may cause larger differences between the measurements and the reconstructed depth map.
To achieve the best results with few iterations, we start with $\lambda=1$ and restart the conjugate gradient optimization procedure five times, while consecutively shrinking the multiplier to a final value of $\lambda=10^{-2}$.

Following the methodology described in the work of comparable depth map SR approaches, we use the Middlebury stereo matching online evaluation tool\footnote{http://vision.middlebury.edu/stereo/eval/} to quantitatively assess the accuracy of our results with respect to the ground truth data. We report the percentage of bad pixels over all pixels in the depth map with an error threshold of $\delta=1$. Additionally, we provide the root-mean-square error (RMSE) based on 8-bit images. We rely on the results reported by the authors of comparable methods regarding the numerical comparison in Table \ref{tab:middlebury_results} and Table \ref{tab:rmse_results}, since an implementation is not publicly available. To show the advantage of enforcing a coupled co-support in the analysis formulation, we further employed a single modal operator learned by the code provided by the authors of \cite{Hawe2012}. This operator has been learned from the same training images as described above, and with the parameters documented in their paper. 

\begin{figure}[hbt]
	\centering
	\begin{subfigure}[b]{0.3\linewidth}
		\centering
		\includegraphics[width=\textwidth]{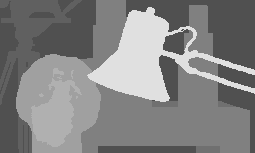}
		\caption{ground truth}
		\label{fig:tsukuba_gt}
	\end{subfigure}%
	\,
	\begin{subfigure}[b]{0.3\linewidth}
		\centering
		\includegraphics[width=\textwidth]{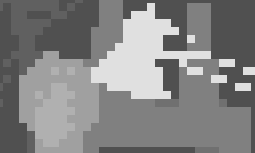}
		\caption{nearest neighbor}
		\label{fig:tsukuba_nn}
	\end{subfigure}
	
	\medskip
	\begin{subfigure}[b]{0.3\linewidth}
		\centering
		\includegraphics[width=\textwidth]{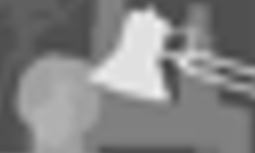}
		\caption{bicubic}
		\label{fig:tsukuba_bc}
	\end{subfigure}%
	\,
	\begin{subfigure}[b]{0.3\linewidth}
		\centering
		\includegraphics[width=\textwidth]{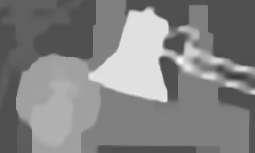}
		\caption{proposed method}
		\label{fig:tsukuba_jid}
	\end{subfigure}
	\caption{Visual comparison of different upscaling methods on a detail in the test image Tsukuba from \cite{Scharsteina} which was downsampled by a factor of 8 in both vertical and horizontal direction.}\label{fig:tsukuba_comparison}
\end{figure}

As illustrated in Figure \ref{fig:tsukuba_comparison}, our method improves depth map SR considerably over simple interpolation approaches. Neither staircasing nor substantial blurring artifacts occur, particularly in areas with discontinuities. Also, there is no noticeable texture cross-talk in areas of smooth depth and cluttered intensity. Edges can be preserved with great detail due to the additional knowledge provided by the intensity image, even if SR is conducted using large upscaling factors. The quantitative comparison with other depth map SR methods demonstrates the superior performance of our JID analysis operator across all test images. It reaches near perfect results for small upscaling factors and the improvement over state-of-the-art methods is of particular significance for larger magnification factors. We refer the reader to the supplementary material for illustrations of our synthetic test results.

%% file: sections/42_kinect.tex
\begin{figure*}[hbt]
	\centering
	\begin{subfigure}[b]{0.32\textwidth}
		\centering
		\includegraphics[width=\textwidth]{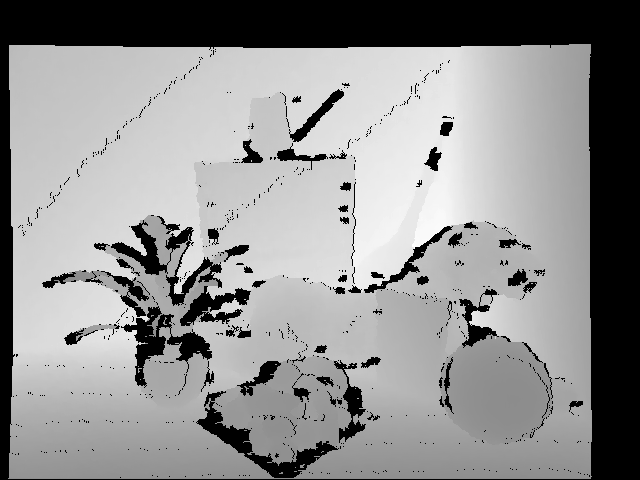}
	\end{subfigure}%
	\,
	\begin{subfigure}[b]{0.32\textwidth}
		\centering
		\includegraphics[width=\textwidth]{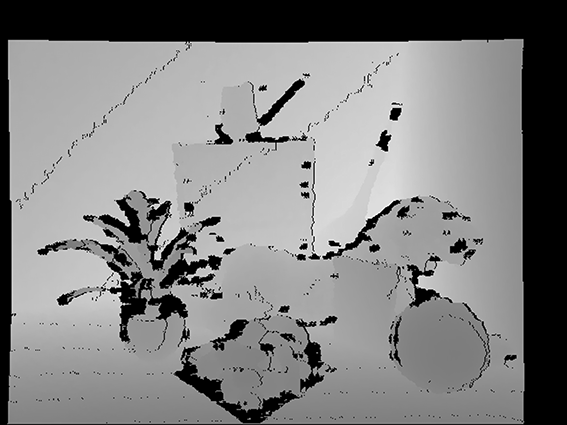}
	\end{subfigure}%
	\,
	\begin{subfigure}[b]{0.32\textwidth}
		\centering
		\includegraphics[width=\textwidth]{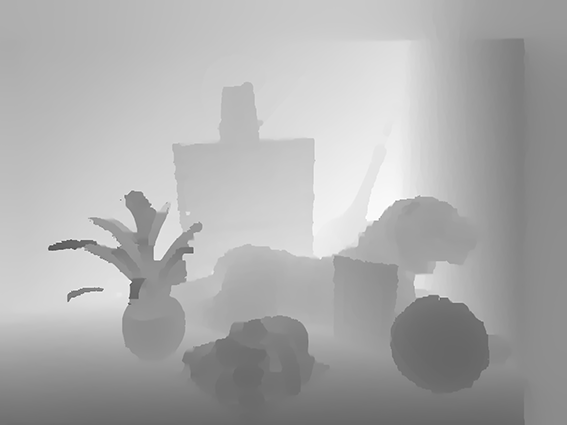}
	\end{subfigure}%
	
	\medskip
	\begin{subfigure}[b]{0.32\textwidth}
		\centering
		\includegraphics[width=\textwidth]{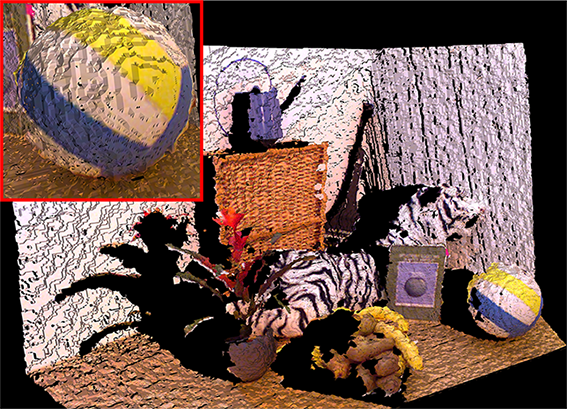}
	\end{subfigure}%
	\,
	\begin{subfigure}[b]{0.32\textwidth}
		\centering
		\includegraphics[width=\textwidth]{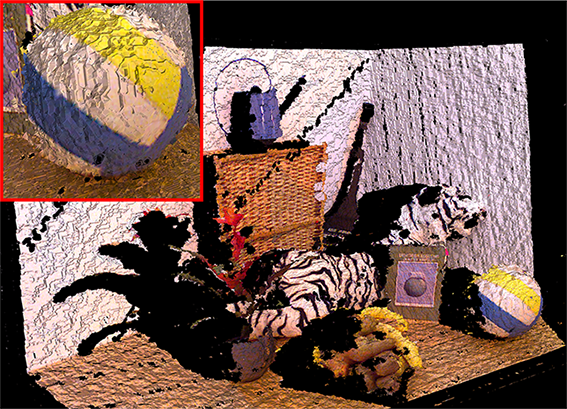}
	\end{subfigure}%
	\,
	\begin{subfigure}[b]{0.32\textwidth}
		\centering
		\includegraphics[width=\textwidth]{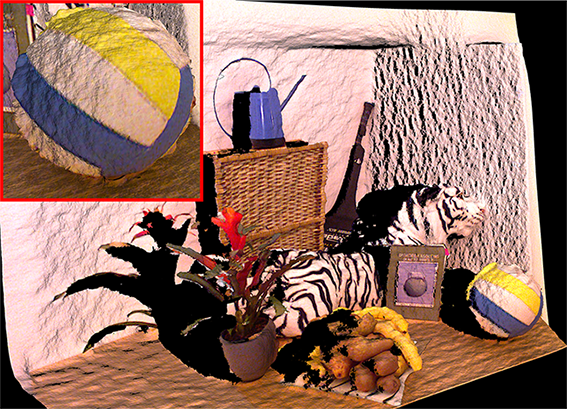}
	\end{subfigure}%

	\medskip
	\begin{subfigure}[b]{0.32\textwidth}
		\centering
		\includegraphics[width=\textwidth]{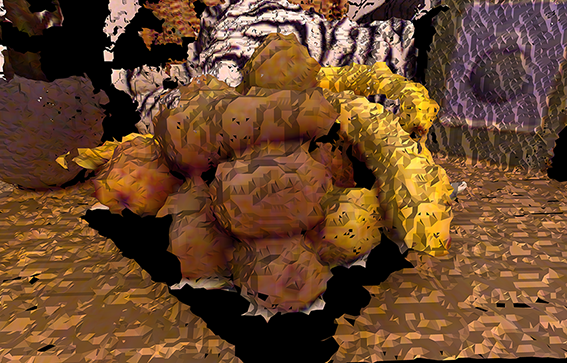}
	\end{subfigure}%
	\,
	\begin{subfigure}[b]{0.32\textwidth}
		\centering
		\includegraphics[width=\textwidth]{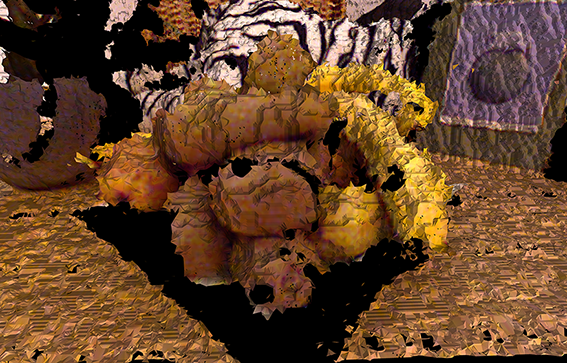}
	\end{subfigure}%
	\,
	\begin{subfigure}[b]{0.32\textwidth}
		\centering
		\includegraphics[width=\textwidth]{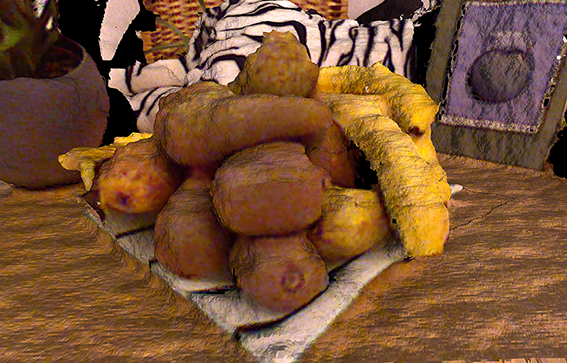}
	\end{subfigure}%
	
	\caption{Depth maps (top row), 3D rendering of Kinect color and depth data depicting the entire scene (middle row) and a detail of the fruit bowl (bottom row). Left column: original Kinect data like in the top row of Figure \ref{fig:eye_catcher} with downsampled color information, center column: bicubic interpolation (1280x960), right column: proposed method (1280x960). Note that object shadows are due to the single view occlusion.}
	\label{fig:kinect_results}
\end{figure*}

In order to demonstrate the applicability of our algorithm to real data, we captured color images of size 1280x960 and corresponding depth maps of size 640x480 using the Microsoft Kinect sensor and then upscale the depth map by a factor of $d=2$ to match its size to the one of the color image.

Since the approximate error statistics for this application and this sensor have been studied previously in \cite{Khoshelham2012}, we can use this information to further refine our data model. According to \cite{Khoshelham2012}, the standard deviation of Kinect depth data is proportional to the square of the depth value $\sigma_i \propto (y_D^{(i)})^{2}$. We utilize this in our error model by employing the squared Mahalanobis distance for $d_{E}$ in \eqref{eq:depth_reconstruction}, which yields
%
\begin{equation}
\mathbf{s}^\star_D \in \underset{\mathbf{s}_D\in \mathbb{R}^{N}} {\arg\min} \, \lambda g(\mathbf{c},\AO^{F}_D\mathbf{s}_D) + d_{E}\left(\mathcal{A}_D\mathbf{s}_D,\mathbf{y}_D \right),
\label{eq:depth_reconstruction_kinect}
\end{equation}
where $d_{E}=\left( \mathcal{A}_D\mathbf{s}_D-\mathbf{y}_D \right)^{\top} \boldsymbol{\Sigma}^{-1} \left( \mathcal{A}_D\mathbf{s}_D-\mathbf{y}_D \right)$ and $\boldsymbol{\Sigma} \in \mathbb{R}^{m_2 \times m_2}$ being a diagonal matrix with main diagonal elements $(y_D^{(i)})^{2}$.

As the Kinect sensor uses structured light to measure depth, the signal is corrupted by missing pixels due to occlusions arising from the displacement of the IR light source and the sensor. To fill these gaps in the data, we model the measurement matrix in such a way that it excludes these gaps from the sampling process of the LR depth image, i.e.\ removing the rows of $\mathcal{A}$ that correspond to zero entries in $\mathbf{y}_D$. As a result, we perform inpainting of missing depth values without any additional processing, while simultaneously increasing the depth map resolution. By this, we handle two of the main issues of Kinect data in one step.

To our knowledge, there is no data set publicly available that allows to numerically evaluate Kinect depth map enhancing methods by providing ground truth data. Therefore, we assess the quality of the super-resolved Kinect depth maps visually. Since small differences in the depth map represented as a gray-scale image are almost invisible to the naked eye, we illustrate our results in Figure \ref{fig:kinect_results} using ball pivoting surface reconstruction \cite{Bernardini1999} on a point cloud that we created from the depth map computed by our algorithm. As it can be seen, our method does not only increase the details in the 3D scene significantly, but also treats the missing pixels with great success. This is especially obvious in the details of the tiger head in Figure \ref{fig:eye_catcher} and the fruit bowl in Figure \ref{fig:kinect_results}. The 3D rendering illustrates the impact of the bimodal support during reconstruction particularly around depth discontinuities, but it also leads to smoother surfaces of table and wall due to the smooth texture of the corresponding intensity signal.

We would like to emphasize that we use the same JID analysis operators as in the Middlebury experiments in Section \ref{sec:middlebury_evaluation}, even though the training data was captured using a different sensor technology than the Kinect. This underpins that the prior model we learn is general enough to be used for high quality reconstruction of both synthetic and real world data.

%% file: sections/50_discussion.tex
\begin{figure}[b]
	\centering
	\newlength\figureheight 
	\newlength\figurewidth
	\newlength\xshift
	\newlength\yshift	
	\setlength\figureheight{3.0cm}
	\setlength\figurewidth{0.5\linewidth}
	\setlength\xshift{43pt}
	\setlength\yshift{-75pt}
	\input{figures/jid_convergence.tikz}
	\caption{Plot of the relative RMSE over the optimization iterations for the upscaling of the synthetic test images by a factor of $8$.}
	\label{fig:reconstruction_convergence}
\end{figure}
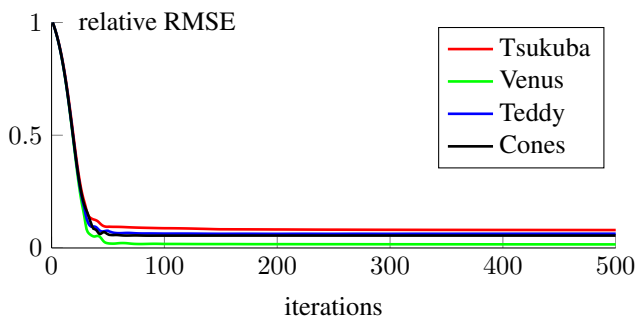

We proposed an approach for inferring high-resolution depth maps from low-resolution depth samples given an additional high-resolution intensity image of the scene. We present an extension of the co-sparse analysis model to the bimodal case. The required pair of analysis operators is learned jointly such that the co-sparse representation of a pair of corresponding intensity and depth samples have a correlated co-support. This data model is employed for depth map super resolution and yields improved results on the benchmark data set over state-of-the-art methods. Moreover, it greatly improves real-world depth data recorded by a Kinect sensor. The fact that the same pre-trained operators can be used to refine both synthetic as well as real-world depth maps, underpins the validity of the model assumptions and emphasizes the capability of this method to abstract training data appropriately.

Despite these compelling results, our method certainly has a few limitations.
%
We showed that missing pixels can be recovered very successfully with our approach. However, the local assumptions fail if the missing areas in the input signal are too large. As a results, inpainting of such large gaps may be inaccurate if the global support in our reconstruction model is insufficient to overcome this.
For instance, this can be observed in the frame of missing pixels around the depth map in Figure \ref{fig:kinect_results}, which is due to registering intensity and depth inputs.
%
%
Finally, in our current implementation, reconstructing a HR depth image with $500$ iterations takes up to three minutes on a single 3.2 GHz CPU with unoptimized Matlab code. Since most of the processing time is dedicated to parallelizable filtering operations, we expect to improve on this with a better software implementation and processing on a GPU. Furthermore, the number of iteration in the reconstruction may be reduced significantly. As shown in Figure \ref{fig:reconstruction_convergence}, the last $400$ iterations only reduce the RMSE by about $0.2$\% and very descent recovery results are achieved with only $50$ optimization steps.


%% file: figures/jid_convergence.tikz
%
%
%
%
\begin{tikzpicture}

\begin{axis}[%
width=\figurewidth,
height=\figureheight,
scale only axis,
xmin=0, xmax=500,
xtick={0,100,200,300,400,500},
xlabel={iterations},
ymin=0, ymax=1,
ytick={0,0.5,1},
ylabel={relative RMSE},
axis x line*=bottom,
axis y line*=left,
legend style={draw=black,fill=white,legend cell align=left},
ylabel style={xshift=\xshift,yshift=\yshift,rotate=-90}
]
\addplot [
color=red,
solid,
line width=1.0pt
]
table{
1 1
2 0.986348220609191
3 0.971315880431964
4 0.955103539841613
5 0.937659224044425
6 0.918281292435942
7 0.897427757368589
8 0.874911009164645
9 0.850565938223576
10 0.824311594577708
11 0.796291695046971
12 0.766329689521122
13 0.73453882280299
14 0.700947518351049
15 0.665720966812742
16 0.629021081313674
17 0.59113916200946
18 0.552276661299381
19 0.512889593932556
20 0.47345522415399
21 0.434490102735307
22 0.39659204962796
23 0.36041322319498
24 0.326578536071817
25 0.295634933716317
26 0.267890640719363
27 0.24357096889259
28 0.222734784385661
29 0.205274426153672
30 0.187103232735532
31 0.166264019819629
32 0.14904706841448
33 0.140174620194048
34 0.135317169316622
35 0.1317485210312
36 0.128958047718655
37 0.126879760408897
38 0.125369828389441
39 0.123762413631261
40 0.121456278008581
41 0.118207806595993
42 0.113628021721837
43 0.108503507946063
44 0.103901657878985
45 0.100063815307721
46 0.0973595950822194
47 0.0955777352392969
48 0.0944576321641015
49 0.0939624632539237
50 0.0940756586891176
51 0.0944310676285897
52 0.0941368872155709
53 0.0939693276040149
54 0.0938112008990482
55 0.0937513335783947
56 0.0937304839878024
57 0.0936884991844826
58 0.0936417546303513
59 0.0935741484112032
60 0.0934696794135052
61 0.0933384537881921
62 0.093187136925817
63 0.0929774021186462
64 0.0927660118008494
65 0.0925320541624069
66 0.0922741325193753
67 0.0920464807731776
68 0.091809430172847
69 0.0915888909298895
70 0.0913680507441105
71 0.0911641974296586
72 0.0909737128582555
73 0.0907959779000675
74 0.0906160612665365
75 0.0904552254849077
76 0.0903158750880685
77 0.0901505172396289
78 0.0900347973743657
79 0.0898901540389922
80 0.0897481475965416
81 0.0896164602248553
82 0.0895103606157979
83 0.0893875665276913
84 0.0892644391874432
85 0.0891593094912377
86 0.0890537213968732
87 0.088935254913756
88 0.0888113644709475
89 0.0887113056146236
90 0.0885892838547729
91 0.0884697468031141
92 0.0883585202998114
93 0.0882406411025491
94 0.0881277546602538
95 0.0880075368085418
96 0.0879099642061937
97 0.0878264765607235
98 0.087736015313758
99 0.0876626439901103
100 0.0875732156314903
101 0.0875573501810948
102 0.0875495050826193
103 0.087532891061602
104 0.087517841586383
105 0.0874924344875359
106 0.0874661321191349
107 0.0874290991036484
108 0.0873850051187008
109 0.0873319177040962
110 0.0872798657520913
111 0.0872201949033546
112 0.0871498795690529
113 0.087092155745831
114 0.0870163177627431
115 0.0869386270821931
116 0.0868727623420823
117 0.0867846776714384
118 0.0867012907985468
119 0.0866141077035035
120 0.0865000412710766
121 0.0863940796218386
122 0.0862971475373796
123 0.0861816202988328
124 0.0860974038111724
125 0.08596635239007
126 0.0858250624238445
127 0.0857088988629578
128 0.0855991504923
129 0.0854662497280693
130 0.0853220786407863
131 0.0851963624624435
132 0.0850639228878004
133 0.0849506597304934
134 0.0848568684086077
135 0.0847333134673453
136 0.0846041828881195
137 0.0844759172519438
138 0.084338161865267
139 0.0842042443980121
140 0.0840499475541719
141 0.083921240970558
142 0.083766670067922
143 0.0836233351965454
144 0.0834983890126165
145 0.083362568732329
146 0.0832534119748359
147 0.0831165840525408
148 0.0830107525408293
149 0.0828947777674097
150 0.0828874699520002
151 0.0828769495222396
152 0.0828637567547028
153 0.0828580187552621
154 0.082855790096618
155 0.0828460989001421
156 0.0828315521663332
157 0.0827908567003633
158 0.0827860621651214
159 0.0827553914159432
160 0.0827439156324122
161 0.0827223310080321
162 0.0827081276625115
163 0.0827010746361181
164 0.0826911357340044
165 0.0826807833973637
166 0.082662486229394
167 0.0826348658048488
168 0.0826219372560786
169 0.0826105108286708
170 0.0825959539448642
171 0.0825826640433224
172 0.0825724600850203
173 0.0825550378174289
174 0.0825467904830868
175 0.0825363782226505
176 0.0825124804634687
177 0.0824925821679509
178 0.0824722239380082
179 0.0824443827762685
180 0.0824320890704656
181 0.08241215042358
182 0.0823918456312784
183 0.0823860066845364
184 0.0823591683908153
185 0.0823356855676667
186 0.0823215150914913
187 0.0823031642639692
188 0.0822824026112195
189 0.0822599049074872
190 0.0822465082472101
191 0.0822218493308737
192 0.0821928840231832
193 0.082177586390313
194 0.0821481170341936
195 0.082135941823313
196 0.082115910252131
197 0.0820818143495623
198 0.0820459899884285
199 0.0820490504435406
200 0.0820482932281618
201 0.0820435552149391
202 0.0820274670660222
203 0.0820160735490742
204 0.0819969808628042
205 0.0819791678870979
206 0.0819660411676322
207 0.0819516119829306
208 0.0819260374999933
209 0.0819101422300637
210 0.0818871102102558
211 0.0818593493875715
212 0.0818353804523678
213 0.0818064207054488
214 0.0817869256349018
215 0.081767541999832
216 0.0817449659328389
217 0.0817208525785865
218 0.0816972654957492
219 0.0816726837518645
220 0.0816495636123376
221 0.0816257339309278
222 0.0815988306394572
223 0.0815802009001989
224 0.081559741888796
225 0.0815357907088381
226 0.0815128011176304
227 0.081498799905095
228 0.0814816828680982
229 0.0814603305057998
230 0.0814413565238003
231 0.0814234113967877
232 0.0813938121051613
233 0.0813698545023105
234 0.0813479583266011
235 0.0813357216891948
236 0.0813216687669791
237 0.0813039943931371
238 0.0812912575480393
239 0.0812757212024711
240 0.0812678324642697
241 0.0812505287071398
242 0.0812363654492544
243 0.0812192196646319
244 0.0811956462695218
245 0.0811884193372881
246 0.0811692489087371
247 0.0811582931044902
248 0.0811540932531997
249 0.0811532692062381
250 0.0811550661494865
251 0.0811520038791311
252 0.0811466977968499
253 0.0811449325703228
254 0.0811348083608328
255 0.0811225023422748
256 0.08111465189281
257 0.0811031088630228
258 0.0810903564266299
259 0.0810742813499365
260 0.0810627325715352
261 0.0810477752565924
262 0.0810332304799362
263 0.0810134960930953
264 0.0809986202384819
265 0.0809873536755989
266 0.0809659721951045
267 0.0809515619525614
268 0.0809441479919558
269 0.0809305761797212
270 0.0809133762619054
271 0.0808990526476711
272 0.0808903537872171
273 0.0808782719662782
274 0.080865740171055
275 0.0808591560658273
276 0.0808462103864041
277 0.0808327769191829
278 0.0808141948131069
279 0.0808051611171079
280 0.0807916246999429
281 0.0807838649340641
282 0.0807708270628025
283 0.0807499866680451
284 0.0807341217113014
285 0.0807184300248145
286 0.0807049806343488
287 0.0806894918146873
288 0.0806809093970839
289 0.0806750107950983
290 0.0806715398839517
291 0.0806591685344412
292 0.0806518293362073
293 0.0806345646091035
294 0.080617954453941
295 0.0806024489110572
296 0.0805853877618322
297 0.0805868440159586
298 0.0805873258589563
299 0.0805883163049596
300 0.0805868600774383
301 0.0805862176157545
302 0.0805830534172282
303 0.080581725594713
304 0.0805670966821629
305 0.0805619877172496
306 0.0805444626840024
307 0.080532795067859
308 0.0805197969173213
309 0.0805114697116726
310 0.0804973479316088
311 0.0804812616262593
312 0.0804643195547766
313 0.0804489292018772
314 0.080433820197322
315 0.0804209401717081
316 0.0804165676823795
317 0.080402134097426
318 0.0803892436280548
319 0.0803785411434695
320 0.0803738550033178
321 0.0803662589036794
322 0.0803631504693291
323 0.080352943866115
324 0.0803465540122461
325 0.0803321722972884
326 0.0803192920863707
327 0.0803111697238749
328 0.0803008812932164
329 0.0802855079752703
330 0.0802744423080293
331 0.0802632514808224
332 0.0802523279007234
333 0.0802437094799668
334 0.0802306860051342
335 0.0802193277233195
336 0.0802190050213657
337 0.0802091781252067
338 0.0801957724480926
339 0.0801925121413184
340 0.0801878258906523
341 0.0801778229767993
342 0.080170595781084
343 0.0801621192746527
344 0.0801530874453345
345 0.0801487272588467
346 0.0801476506330755
347 0.0801466924239754
348 0.08014773138051
349 0.0801435647066812
350 0.0801416697135879
351 0.0801384125895137
352 0.080136689759615
353 0.0801342077426759
354 0.0801302934295929
355 0.0801292004029307
356 0.0801220980586092
357 0.0801125931784273
358 0.0801044498572551
359 0.080095400751586
360 0.0800883546685419
361 0.0800846913223443
362 0.0800746162561392
363 0.0800675414012736
364 0.0800646746408312
365 0.080062804724855
366 0.0800579330383593
367 0.080046755061973
368 0.0800432730783697
369 0.0800337372735588
370 0.0800257207123795
371 0.080016107327622
372 0.0800074742463336
373 0.0799991476439918
374 0.0799980797916165
375 0.0799974865341377
376 0.0799923250084198
377 0.0799836246297464
378 0.0799755437031739
379 0.079967607632828
380 0.0799630269047217
381 0.0799587480841952
382 0.0799506646432529
383 0.0799402326667808
384 0.0799337720380967
385 0.0799201906013628
386 0.0799183982839191
387 0.0799144032148441
388 0.0799047440402515
389 0.0798969737393826
390 0.0798929775989678
391 0.0798829971753389
392 0.0798768019886715
393 0.0798704550698821
394 0.0798607149447003
395 0.079861076911813
396 0.0798591698129043
397 0.0798601638871398
398 0.0798587105896495
399 0.0798583432091444
400 0.0798547287431818
401 0.0798517516834762
402 0.0798517678928133
403 0.0798524540850624
404 0.0798478667508051
405 0.0798406907453222
406 0.0798402692430186
407 0.0798391992656386
408 0.0798343247427736
409 0.079828287920001
410 0.0798211750165777
411 0.079815012870523
412 0.0798074446717941
413 0.0798037305860452
414 0.0798003028775564
415 0.0798020437826012
416 0.0798037900560288
417 0.07980529841625
418 0.0798049416024279
419 0.0798021951638538
420 0.0797984754270554
421 0.0797944148768778
422 0.079790581225096
423 0.0797810206252669
424 0.0797771484726812
425 0.0797725081316553
426 0.0797638162304868
427 0.0797555399247171
428 0.0797467704309974
429 0.0797379458645924
430 0.0797322534740638
431 0.0797238223619203
432 0.07972058063798
433 0.0797176310417302
434 0.0797122024169444
435 0.0797061455188479
436 0.0797021289840943
437 0.0796928501231756
438 0.079687181582629
439 0.0796846693210952
440 0.0796770237620565
441 0.0796719335355019
442 0.0796672112473094
443 0.0796626944850072
444 0.0796621095618074
445 0.0796605118330311
446 0.0796602464444699
447 0.0796586324296909
448 0.0796525606248055
449 0.0796514989592857
450 0.0796520947937177
451 0.0796498576832106
452 0.0796486226412901
453 0.079645599957905
454 0.0796425229867442
455 0.0796423550500379
456 0.0796363307636692
457 0.0796314492405662
458 0.0796229966172049
459 0.0796202655778698
460 0.0796156594461277
461 0.0796141745892154
462 0.0796117304761064
463 0.0796087334899916
464 0.0796003109755697
465 0.0795982458601291
466 0.0795985222960105
467 0.0795974870115981
468 0.0795952808894233
469 0.0795912533391426
470 0.0795847264319137
471 0.0795781989893558
472 0.0795737097064981
473 0.0795727988066048
474 0.0795712263953483
475 0.0795667746763478
476 0.0795612489929254
477 0.0795553975319614
478 0.079549263610844
479 0.0795465951261733
480 0.0795388061101561
481 0.0795289765417178
482 0.0795279078012857
483 0.079519476725999
484 0.0795172630136708
485 0.07951387179071
486 0.0795090315867721
487 0.0795058896374924
488 0.0795008101615289
489 0.0794970437593478
490 0.0794920776969831
491 0.0794892064658395
492 0.0794892064658395
493 0.0794892064658395
494 0.0794892064658395
495 0.0794892064658395
496 0.0794892064658395
497 0.0794892064658395
498 0.0794892064658395
499 0.0794892064658395
500 0.0794892064658395
};
\addlegendentry{Tsukuba};

\addplot [
color=green,
solid,
line width=1.0pt
]
table{
1 1
2 0.985748692375818
3 0.970224246031579
4 0.952981076721831
5 0.934407246109918
6 0.9140852791712
7 0.892099696401174
8 0.868339584941995
9 0.842727766238784
10 0.815162206565555
11 0.785656765527555
12 0.754216562521327
13 0.720836459494733
14 0.685620422202988
15 0.648643118754074
16 0.61011348342975
17 0.570259583675607
18 0.529376729178299
19 0.487813703935246
20 0.446013671477301
21 0.404498492459273
22 0.363819091405624
23 0.32453537219869
24 0.287334206371288
25 0.252675882840305
26 0.221080121019236
27 0.192796536958206
28 0.168125414454924
29 0.141057484397091
30 0.109517462121298
31 0.0833434677538643
32 0.0711722462742119
33 0.0646574222483669
34 0.0603796905762549
35 0.056665323865479
36 0.0537983069573952
37 0.0515146275494279
38 0.0507548216156771
39 0.0517155369099075
40 0.0530999063094912
41 0.0534670486436022
42 0.052101166793172
43 0.0485156672740389
44 0.0430496965475329
45 0.0363940502801494
46 0.0301798090954085
47 0.0257210987938417
48 0.0231203969893536
49 0.0216004735931647
50 0.0207252748477131
51 0.0199463726410387
52 0.0195667129929967
53 0.0193964294348417
54 0.0192238728015749
55 0.0192532312152475
56 0.0194290502928542
57 0.0198565945797267
58 0.0203730151325976
59 0.0209150989141099
60 0.0213155650817373
61 0.0216387934167573
62 0.0217801814402565
63 0.0218209508027938
64 0.0217725912692024
65 0.0216007787315109
66 0.021312139610575
67 0.0208902123081214
68 0.0203965193517614
69 0.0198398117979023
70 0.0193217618802777
71 0.0188963104793647
72 0.0184886374575343
73 0.0181441489649722
74 0.0179035800247542
75 0.0177240141511345
76 0.0176104562817396
77 0.0175039259215457
78 0.0174551666358725
79 0.017443515047776
80 0.0174771411041008
81 0.0175445177584759
82 0.0176377285557852
83 0.0177697248649383
84 0.0178947422751575
85 0.0180094918084197
86 0.0181445401718914
87 0.0182598857144486
88 0.0183789767677596
89 0.0184602328392467
90 0.0185189150741739
91 0.0185286044075577
92 0.0185160401443555
93 0.0184828228684581
94 0.0184133451451965
95 0.0183138673694246
96 0.0181972480383502
97 0.0181031092132469
98 0.0179971849226528
99 0.0178998698343962
100 0.0177931060919792
101 0.0177789099401507
102 0.0177806494353568
103 0.017773034321021
104 0.0177684123237393
105 0.0177629043002911
106 0.0177574231205933
107 0.0177541393120249
108 0.01775096914673
109 0.017751683197318
110 0.017742741211602
111 0.0177333944488793
112 0.0177118237873954
113 0.0176945254022433
114 0.0176839776932979
115 0.01765235419573
116 0.0176167890525242
117 0.017592107175533
118 0.0175558712748993
119 0.0175348918499543
120 0.0175075752216992
121 0.0174812600544299
122 0.017459958672306
123 0.0174447357784754
124 0.0174141044939796
125 0.0173900096178605
126 0.0173737332520494
127 0.0173602455733268
128 0.0173412516397146
129 0.0173201000714035
130 0.0173031133175655
131 0.0172901277108525
132 0.0172874297025091
133 0.0172785702356022
134 0.0172732875973341
135 0.0172747258088858
136 0.017268003342998
137 0.0172603970495368
138 0.0172516706470144
139 0.017253522070908
140 0.017253257593944
141 0.0172475556601591
142 0.0172417048101291
143 0.0172260241510772
144 0.0172199598829294
145 0.0172118905065567
146 0.0172090329226644
147 0.0172009289442679
148 0.0171946789064643
149 0.0171935288879378
150 0.0172001920296249
151 0.017204583574358
152 0.0172159551177763
153 0.0172262596135732
154 0.0172355872258722
155 0.0172442335617584
156 0.0172578706663022
157 0.0172581056942975
158 0.0172525816899497
159 0.0172402638439108
160 0.0172102702852738
161 0.0171921133749162
162 0.017183293282901
163 0.017167382071082
164 0.0171443008497211
165 0.017124386443223
166 0.0171045970611964
167 0.0170727022162662
168 0.0170415807388067
169 0.0170150812375686
170 0.0169863616160345
171 0.0169601644671615
172 0.016942517595515
173 0.0169208076778469
174 0.0168968495414168
175 0.0168883856139226
176 0.0168738790240412
177 0.0168703330750228
178 0.016847868098117
179 0.0168486806069945
180 0.0168345012162831
181 0.0168279041819994
182 0.0168169335229506
183 0.0168281150861198
184 0.0168304348570148
185 0.0168302541071331
186 0.0168349830900567
187 0.016847868098117
188 0.0168438952683014
189 0.016836699653293
190 0.0168373621396644
191 0.0168326338251142
192 0.0168398612854227
193 0.0168357661055228
194 0.0168380848223423
195 0.0168394096600329
196 0.0168429621193594
197 0.016846844883047
198 0.0168514789488193
199 0.0168643814176252
200 0.0168688303294166
201 0.0168761624611435
202 0.016877003649566
203 0.016885563361966
204 0.0168798273326786
205 0.0168796170746822
206 0.0168818396695874
207 0.0168745100046606
208 0.0168675979781179
209 0.0168604124788541
210 0.0168538857528404
211 0.0168476273472207
212 0.016841276299885
213 0.0168308264750993
214 0.0168171144160016
215 0.0168098168521792
216 0.0168004942807678
217 0.0167900794685175
218 0.0167754274216003
219 0.0167715583547177
220 0.0167575860104773
221 0.016751321893724
222 0.0167423604375294
223 0.0167349393862151
224 0.0167320609501378
225 0.0167220583222884
226 0.0167155989021357
227 0.0167103203397243
228 0.0166981793863794
229 0.0166948694457986
230 0.0166962663814835
231 0.0166906171830009
232 0.0166796474180076
233 0.0166768810476733
234 0.0166767898408378
235 0.0166770938616829
236 0.0166704041223154
237 0.0166670886658067
238 0.0166732019762163
239 0.0166761209754686
240 0.016675543297423
241 0.0166787962760054
242 0.0166752392483082
243 0.0166701608084799
244 0.0166647461567627
245 0.0166626467455066
246 0.0166624033183935
247 0.0166654154786595
248 0.0166690658522694
249 0.0166727458364192
250 0.0166777018867471
251 0.0166786746807451
252 0.0166717726964746
253 0.0166746919459303
254 0.0166757257242309
255 0.0166827476395033
256 0.0166766074256703
257 0.0166785530845983
258 0.0166718335193852
259 0.0166702520515844
260 0.0166759993607009
261 0.0166697654159227
262 0.0166662368824311
263 0.016656286044583
264 0.0166473952539828
265 0.0166418513309283
266 0.0166329833070219
267 0.0166281968763857
268 0.0166189249643924
269 0.0166093426212013
270 0.0165972194464835
271 0.0165918421068509
272 0.0165871966379133
273 0.0165748125518917
274 0.0165724264058777
275 0.0165621743057838
276 0.0165539986821403
277 0.0165458190187564
278 0.0165406088857372
279 0.0165399651671795
280 0.0165395973167552
281 0.0165386776548997
282 0.0165365009180812
283 0.0165382178047942
284 0.0165260424544138
285 0.0165246004462528
286 0.0165226366583079
287 0.0165202429761291
288 0.0165171429491794
289 0.0165136125069784
290 0.0165059043104348
291 0.0165016648038202
292 0.0164975778715544
293 0.0164918913567736
294 0.0164867564401214
295 0.0164810969549692
296 0.0164779895634261
297 0.0164797433118687
298 0.0164813738228164
299 0.0164824812476976
300 0.0164823889651327
301 0.0164852187287983
302 0.0164862643881028
303 0.0164811892447684
304 0.0164824812476976
305 0.0164830964515933
306 0.0164831887301972
307 0.0164769741500638
308 0.0164731580979651
309 0.0164680173400691
310 0.0164654617555276
311 0.0164576385856103
312 0.0164517533577622
313 0.0164492568923264
314 0.0164544959509631
315 0.0164484863017838
316 0.016444324488867
317 0.0164344243902353
318 0.0164318944366095
319 0.0164241787581013
320 0.0164188991458004
321 0.0164158417520525
322 0.0164118261275694
323 0.0164070060808587
324 0.0164025864604271
325 0.0164036992042975
326 0.0163995569418154
327 0.0163994332761068
328 0.0163928776585484
329 0.0163909908866142
330 0.0163820489643756
331 0.0163755483107296
332 0.0163678989996273
333 0.0163608349228026
334 0.016360463044795
335 0.0163604320545795
336 0.0163510393141343
337 0.0163502640945822
338 0.0163494268161798
339 0.0163445263271287
340 0.0163410206456143
341 0.0163386003558529
342 0.0163362417798997
343 0.0163319271874497
344 0.0163317719650441
345 0.0163290708589853
346 0.0163273940856414
347 0.0163257792522468
348 0.0163232013987152
349 0.0163237604858581
350 0.0163235741256048
351 0.0163238847248449
352 0.01632248698172
353 0.0163238536651869
354 0.016316304415454
355 0.016312948071943
356 0.0163065442419328
357 0.0163040566395855
358 0.0163006666703962
359 0.0162964048832669
360 0.0162890919217562
361 0.0162869752182294
362 0.016282118211553
363 0.0162750169021166
364 0.0162695953964229
365 0.0162639850404776
366 0.0162596200919365
367 0.0162568758101327
368 0.016251760230237
369 0.0162502314823248
370 0.0162436468629242
371 0.0162411496267995
372 0.0162359666364508
373 0.0162315004443334
374 0.0162231581792058
375 0.0162170940662879
376 0.0162165938320404
377 0.0162114968160253
378 0.0162008911379148
379 0.0162025184183605
380 0.0162029565044034
381 0.0162010789093848
382 0.0161962900568819
383 0.0161944743047238
384 0.0161876164386199
385 0.0161816330045503
386 0.0161791261966353
387 0.0161782800612948
388 0.0161724499260228
389 0.0161671194688233
390 0.0161717288492949
391 0.0161675271540793
392 0.0161678093916959
393 0.0161697849170748
394 0.0161590577333041
395 0.016159904875129
396 0.0161583046810631
397 0.0161585870797662
398 0.0161560453137361
399 0.0161539739494179
400 0.0161540994942101
401 0.0161503641192847
402 0.0161465022773279
403 0.016144115636814
404 0.0161382417423727
405 0.016135602505565
406 0.0161298512444597
407 0.0161285938650816
408 0.0161265504144971
409 0.0161211733305369
410 0.0161199467268709
411 0.0161139067013445
412 0.0161091863556466
413 0.016101567947373
414 0.0160963085133247
415 0.0160881482654157
416 0.0160829159678788
417 0.01607843879585
418 0.0160739603767689
419 0.0160720046193149
420 0.016071720699258
421 0.016071026651336
422 0.0160687550121776
423 0.0160646210670196
424 0.0160640529608903
425 0.0160610858579584
426 0.0160592232473687
427 0.0160554973779407
428 0.0160546447269348
429 0.0160493383250764
430 0.016049085595511
431 0.0160445357825927
432 0.0160315743820277
433 0.0160331239787653
434 0.016032396635683
435 0.0160302144084246
436 0.0160310683588159
437 0.0160254061348284
438 0.0160202482904861
439 0.0160215774643964
440 0.0160199634531575
441 0.016017019837489
442 0.0160107826491724
443 0.0160132841484421
444 0.0160117959617889
445 0.0160112893134969
446 0.0160096742650607
447 0.016007710653391
448 0.0160046063887478
449 0.0160034658954072
450 0.015995099792792
451 0.0159912004548643
452 0.0159862535849841
453 0.0159767995374927
454 0.0159757205322052
455 0.0159717212299309
456 0.0159646406440539
457 0.0159596220047457
458 0.0159555868908127
459 0.0159537437433971
460 0.0159495164114887
461 0.0159392771754545
462 0.0159367640597537
463 0.0159341232707101
464 0.0159303999691787
465 0.0159267712996893
466 0.0159204669080257
467 0.0159167085577946
468 0.015911897848701
469 0.0159083287009439
470 0.0159075637793814
471 0.0159040255386005
472 0.0159011880020594
473 0.015900390850135
474 0.0159003270762546
475 0.0158963088059894
476 0.0158940759946588
477 0.0158878543656372
478 0.0158866097474458
479 0.0158841840574423
480 0.015881853769524
481 0.0158801617023948
482 0.0158779585523584
483 0.0158774795662455
484 0.0158767131584076
485 0.0158747650386167
486 0.0158741581983755
487 0.0158709639199366
488 0.0158681204713195
489 0.0158662991185597
490 0.0158653723851494
491 0.0158672897047648
492 0.0158672897047648
493 0.0158672897047648
494 0.0158672897047648
495 0.0158672897047648
496 0.0158672897047648
497 0.0158672897047648
498 0.0158672897047648
499 0.0158672897047648
500 0.0158672897047648
};
\addlegendentry{Venus};

\addplot [
color=blue,
solid,
line width=1.0pt
]
table{
1 1
2 0.985720419938235
3 0.970330647343031
4 0.953404341336996
5 0.934972815569493
6 0.914977806708037
7 0.893296443549177
8 0.869816447489048
9 0.844584420240593
10 0.81743885897507
11 0.788402483741022
12 0.757495675719413
13 0.724717154002898
14 0.690193138732651
15 0.654015585552589
16 0.616379797643875
17 0.577508927332468
18 0.537771759591426
19 0.49751942307186
20 0.457198905368122
21 0.417345403581886
22 0.378516114589168
23 0.341336999016926
24 0.306379595792664
25 0.274236759963064
26 0.245240177050274
27 0.219641134050364
28 0.197457704565952
29 0.178733801393872
30 0.152520961890559
31 0.131221646099112
32 0.116185855804908
33 0.105114250186663
34 0.0985402168244923
35 0.0947080836657379
36 0.0943022308589722
37 0.095015149139084
38 0.0950107438158544
39 0.0935763008630084
40 0.0902842314748219
41 0.0850658594873672
42 0.0798959130231221
43 0.0759314962153455
44 0.0738180703483949
45 0.0731571144449675
46 0.0736152293251214
47 0.0745951849396879
48 0.07576822492291
49 0.0765312132168119
50 0.0764099701564781
51 0.0734139159374867
52 0.0719323189505458
53 0.0703230456405626
54 0.0686668395708715
55 0.0675820405668825
56 0.0667899579777895
57 0.0662093976888218
58 0.0658513273708942
59 0.0656364960480236
60 0.0655517343991687
61 0.0656527601648832
62 0.0657939292070366
63 0.0660124516828041
64 0.0662755432485482
65 0.066539026948065
66 0.0667156141687357
67 0.0667963475408077
68 0.0667788151981174
69 0.0666855577353706
70 0.0664912366758372
71 0.0662260280039643
72 0.0659024492691538
73 0.0655741594361594
74 0.065283424730309
75 0.0650154267949943
76 0.0647851538787863
77 0.0646158035457659
78 0.0644899522862153
79 0.0643952028418963
80 0.0643202626791521
81 0.0642841164896324
82 0.0642660923896269
83 0.0642741755543856
84 0.0642944027531262
85 0.0642966409020028
86 0.0643138862748598
87 0.0643118020353973
88 0.0643050816369329
89 0.0642709248782964
90 0.0642433276857422
91 0.0642103116584352
92 0.0641709831479258
93 0.0641056249229432
94 0.0640425056764124
95 0.0639836096233894
96 0.0639290806306248
97 0.0638761589082916
98 0.0638253339572637
99 0.0637731322949879
100 0.0637218346658189
101 0.0637215657324282
102 0.0637259532573263
103 0.0637262995310645
104 0.0637307161980551
105 0.0637395928097459
106 0.0637537670425919
107 0.0637600779628711
108 0.063776437790376
109 0.0637817674488457
110 0.0637903131124882
111 0.0637994868337465
112 0.0637970251824508
113 0.0637961972514771
114 0.0637986626142934
115 0.0637973968279715
116 0.0637874021213634
117 0.0637812264057624
118 0.0637672681159968
119 0.0637532331262769
120 0.0637451428263529
121 0.0637282518903343
122 0.0637095952049007
123 0.063682731624325
124 0.0636483072719557
125 0.063620794229459
126 0.0635945760468214
127 0.063558356953847
128 0.063531787842905
129 0.063483978316416
130 0.0634490169496306
131 0.0634162018782278
132 0.0633757323616305
133 0.0633368418795971
134 0.0632932098823347
135 0.0632570149233621
136 0.0632314255495192
137 0.0631928808956398
138 0.0631416137692329
139 0.0631031519014222
140 0.0630779690335476
141 0.0630644953458351
142 0.0630336364552869
143 0.0630157277983355
144 0.0629898950571531
145 0.0629599765000695
146 0.0629297079852998
147 0.0629080755024018
148 0.0628924714695426
149 0.0628746607495319
150 0.0628766358191957
151 0.0628789020325274
152 0.0628837888604124
153 0.0628926245056426
154 0.0628982044700855
155 0.0629122063215859
156 0.0629281486655903
157 0.0629402193287217
158 0.0629461120748002
159 0.0629579965874903
160 0.0629687529899174
161 0.0629807264143582
162 0.0629910839012955
163 0.0630016967883405
164 0.0630217289572493
165 0.0630317519651507
166 0.0630362880535074
167 0.0630419930805059
168 0.0630473401436135
169 0.0630418701969838
170 0.0630461970421896
171 0.0630484981177353
172 0.0630501289262508
173 0.0630462603415076
174 0.0630298487963894
175 0.0630258560520904
176 0.063017419057998
177 0.0630009888213363
178 0.0629938341820001
179 0.0629861718546227
180 0.0629583881005662
181 0.0629385334542998
182 0.0629208439693321
183 0.0628953156405982
184 0.0628674245200282
185 0.0628446014177577
186 0.0628221810697692
187 0.0628026422047289
188 0.0627877373042385
189 0.0627688646506134
190 0.0627521785483354
191 0.0627370408907054
192 0.0627157836392467
193 0.0627038195245906
194 0.062690407718796
195 0.0626824274187431
196 0.0626735771226984
197 0.0626670031993232
198 0.0626550747307718
199 0.0626601251259206
200 0.0626627775443114
201 0.0626703632907593
202 0.0626835546837
203 0.0626990759223142
204 0.0627176663943585
205 0.0627294856611916
206 0.0627451339682226
207 0.0627662241955259
208 0.0627828953475079
209 0.0628065145820969
210 0.0628232273562817
211 0.0628399207427199
212 0.0628551942202058
213 0.0628685297978272
214 0.0628854052752793
215 0.0628972714001326
216 0.062906795527773
217 0.0629123667725046
218 0.0629206462308323
219 0.0629297005245319
220 0.0629399619747062
221 0.0629461941218009
222 0.0629535928298521
223 0.0629567400005041
224 0.0629571538939432
225 0.0629536897829556
226 0.0629489612807421
227 0.062942632437646
228 0.0629320245109461
229 0.0629278576877781
230 0.0629066126721285
231 0.0628956552902244
232 0.062886181736932
233 0.0628641459127861
234 0.0628515265336909
235 0.0628404960400078
236 0.0628263773208309
237 0.062817823846405
238 0.0628061370731853
239 0.0627867390317474
240 0.0627746949505174
241 0.0627565590328743
242 0.0627379164751122
243 0.0627233854310829
244 0.0627177674551618
245 0.0627097831499124
246 0.0627019588173023
247 0.0626928940993191
248 0.0626960318906166
249 0.0626992294304534
250 0.0627032317417354
251 0.0627076980045383
252 0.0627200693515838
253 0.0627317497060796
254 0.0627443295712727
255 0.0627657117997292
256 0.0627858828228466
257 0.0628020104911743
258 0.0628180293826008
259 0.0628341226474535
260 0.0628501707053697
261 0.0628642056611448
262 0.0628883542854238
263 0.062900701297202
264 0.0629142324508187
265 0.0629300847529265
266 0.0629331435749147
267 0.0629426958411154
268 0.0629520527470702
269 0.0629583508137115
270 0.0629700988090264
271 0.0629697297368424
272 0.062974695254092
273 0.0629749673769789
274 0.0629768312007603
275 0.0629699496892125
276 0.0629679365371603
277 0.0629664825539826
278 0.0629588877422888
279 0.0629502926063257
280 0.0629410212221674
281 0.0629367057960588
282 0.0629256342744347
283 0.0629145085665092
284 0.06289858889088
285 0.0628809217683249
286 0.0628751722574586
287 0.062857240860582
288 0.06283723470711
289 0.0628177976872045
290 0.0628087010989833
291 0.0627959359980539
292 0.0627885598389338
293 0.0627816876043611
294 0.0627759252653321
295 0.0627659212466917
296 0.0627564131463387
297 0.0627574268637181
298 0.0627590876789425
299 0.0627637818626835
300 0.0627685804143242
301 0.0627731504786371
302 0.0627862268033715
303 0.0627986238001007
304 0.0628111192606408
305 0.0628197184909553
306 0.0628346494294927
307 0.0628473095501726
308 0.0628662557422982
309 0.0628838224583729
310 0.0628960733180158
311 0.0629078553336934
312 0.0629220490414734
313 0.0629325616641344
314 0.062941166680814
315 0.0629506506050404
316 0.0629616618003281
317 0.06297493755535
318 0.0629784638650693
319 0.0629834473297405
320 0.0629860749515183
321 0.0629934093497329
322 0.0629980786207684
323 0.0630067193995008
324 0.0630085413011749
325 0.0630143233477751
326 0.0630074422052422
327 0.0630118012187201
328 0.0630055345762771
329 0.0630001914173944
330 0.0630086456211486
331 0.0630125798483583
332 0.0630034294108699
333 0.0630011453199906
334 0.0629909422846752
335 0.0629871856012392
336 0.0629826087039694
337 0.0629756569982071
338 0.0629740801774808
339 0.0629728798288529
340 0.0629652857887922
341 0.062959480595607
342 0.0629530894171071
343 0.0629454855305413
344 0.0629412860312473
345 0.0629304354059189
346 0.0629317335510548
347 0.062932490790018
348 0.0629348072132339
349 0.062937936673771
350 0.0629429009695493
351 0.062948021504554
352 0.0629485622505031
353 0.0629553677937819
354 0.0629566355945222
355 0.0629652112232301
356 0.0629740876329909
357 0.0629817029755942
358 0.0629911696164637
359 0.0629972886327542
360 0.0630060412958543
361 0.063010989049324
362 0.0630174972869722
363 0.0630210547400817
364 0.0630319865984162
365 0.0630360534362528
366 0.0630392374530772
367 0.0630457204335193
368 0.0630482933328631
369 0.0630483119496971
370 0.0630501028634327
371 0.0630513017419155
372 0.0630548423959116
373 0.0630567187502364
374 0.0630618002628405
375 0.0630637248020957
376 0.063065760952802
377 0.0630635312340728
378 0.063062250692186
379 0.0630674322595569
380 0.0630692598516589
381 0.0630728813797608
382 0.0630731530790333
383 0.0630743105792583
384 0.0630740947126605
385 0.0630705774717656
386 0.0630712176614291
387 0.0630725315186997
388 0.06306990749888
389 0.0630708677911392
390 0.0630715638077847
391 0.0630754122317689
392 0.063074072381591
393 0.0630750363322302
394 0.0630727138933247
395 0.0630753936229335
396 0.0630758513986894
397 0.0630754271188332
398 0.0630772209843639
399 0.0630777903957429
400 0.0630793162440306
401 0.0630829223095131
402 0.06308439965762
403 0.0630847234043467
404 0.0630870900540901
405 0.0630909152058956
406 0.0630962655641939
407 0.0630991823973707
408 0.063102102815912
409 0.0631048222178924
410 0.0631088918016695
411 0.0631113170595212
412 0.0631174876558186
413 0.0631209985636965
414 0.0631241001990784
415 0.0631243084569054
416 0.0631298456317769
417 0.0631319354264719
418 0.0631324708787481
419 0.0631337128130731
420 0.0631336198547629
421 0.0631377954070481
422 0.063142524639439
423 0.0631458705804716
424 0.0631477814050342
425 0.0631497404978135
426 0.0631509820925122
427 0.0631544725528208
428 0.0631550635701325
429 0.06315576609295
430 0.0631561638140437
431 0.0631592562888735
432 0.0631637646377455
433 0.0631661134603578
434 0.0631696253818754
435 0.0631681165802574
436 0.0631715317717948
437 0.0631728844190286
438 0.0631735570158002
439 0.0631751400053562
440 0.0631781535213829
441 0.0631797735519713
442 0.0631824933264283
443 0.063184588810502
444 0.0631864538814891
445 0.063184187552974
446 0.0631874346935458
447 0.0631900649779927
448 0.0631917367120472
449 0.063193192942115
450 0.0631950280414005
451 0.0631923125213148
452 0.063193074067449
453 0.0631955592444061
454 0.0631982077633902
455 0.0632001541497812
456 0.063203697607743
457 0.0632072780070863
458 0.063210260280344
459 0.0632121320213296
460 0.0632141708189813
461 0.0632147130018237
462 0.0632160981464629
463 0.0632164212187295
464 0.0632179771413498
465 0.0632207324044032
466 0.0632247388327817
467 0.0632267289567624
468 0.0632288898040321
469 0.063232041835404
470 0.0632358470820718
471 0.0632398599770689
472 0.063241938712469
473 0.0632416788742807
474 0.0632415341068269
475 0.0632428407138945
476 0.0632425066407925
477 0.0632416825862623
478 0.0632452571234532
479 0.0632470461721237
480 0.0632473319698167
481 0.0632469385336967
482 0.0632473245465163
483 0.063247881291626
484 0.0632487572273445
485 0.0632515705320519
486 0.0632513923845414
487 0.0632519936303785
488 0.0632533965151068
489 0.0632542686632084
490 0.0632549589506273
491 0.0632560277678978
492 0.0632560277678978
493 0.0632560277678978
494 0.0632560277678978
495 0.0632560277678978
496 0.0632560277678978
497 0.0632560277678978
498 0.0632560277678978
499 0.0632560277678978
500 0.0632560277678978
};
\addlegendentry{Teddy};

\addplot [
color=black,
solid,
line width=1.0pt
]
table{
1 1
2 0.985822461105112
3 0.970362619015489
4 0.9534660087763
5 0.935042522721838
6 0.9150678337544
7 0.893400196674928
8 0.86996085463694
9 0.844713967046891
10 0.817578472893237
11 0.788581470603255
12 0.757674204272341
13 0.724959830924344
14 0.690469455483332
15 0.654348888446216
16 0.616801923198707
17 0.578093016367367
18 0.538561885701286
19 0.498590044530685
20 0.4587100927592
21 0.41944992118925
22 0.381430747884291
23 0.345306023994776
24 0.311751249751702
25 0.281206840706196
26 0.253904231112799
27 0.229891016843861
28 0.209228617311418
29 0.191764569689275
30 0.177174383859136
31 0.165077346006457
32 0.15313177452415
33 0.13841005946931
34 0.123004433516486
35 0.109636223807211
36 0.0955349545698601
37 0.0859312152114684
38 0.0836963489846026
39 0.0817635740702427
40 0.077186996135555
41 0.0701969331132754
42 0.0638031437697979
43 0.0619515725056362
44 0.0640591246498255
45 0.0677985341863051
46 0.0701301079353493
47 0.0698403392179847
48 0.0670165183773926
49 0.06335993593618
50 0.0606456612197528
51 0.0584159951474557
52 0.0578826476893303
53 0.0574353124632594
54 0.0572312097123215
55 0.0572658522763126
56 0.0575766456199327
57 0.0576896124314365
58 0.0575933534511559
59 0.0572564043900463
60 0.0567960108549782
61 0.0563093170525973
62 0.0558776730033529
63 0.0555287462621746
64 0.0552959806281475
65 0.0551586915207521
66 0.0550837664272088
67 0.0550913711774696
68 0.0551445779099079
69 0.0552418404271968
70 0.0553598014220149
71 0.0554810835020596
72 0.0555852463090467
73 0.0556456620201612
74 0.0556685895516365
75 0.0556437446216648
76 0.0555759587850619
77 0.0554837813668657
78 0.0553570805088418
79 0.0552275690825327
80 0.0550963179739139
81 0.0549832423434192
82 0.0548883830571171
83 0.0548119377442992
84 0.0547604679408469
85 0.0547357984467061
86 0.0547234767322607
87 0.054718043173161
88 0.0547173406419623
89 0.0547322782399207
90 0.0547437008876709
91 0.0547563905605568
92 0.0547676964500436
93 0.0547845907661437
94 0.0547982129908553
95 0.0548035756159052
96 0.054805264748433
97 0.0547990862784355
98 0.0547963146160734
99 0.0547904129133436
100 0.0547766655803546
101 0.0547766884954466
102 0.0547769491539434
103 0.0547743625647129
104 0.0547724862801423
105 0.0547708849421696
106 0.0547686848165112
107 0.0547697619724075
108 0.0547705354489461
109 0.0547700226638686
110 0.0547700341228055
111 0.05477192194992
112 0.0547838604482285
113 0.0547964806907005
114 0.0547991721748228
115 0.0548123499490609
116 0.0548168553739999
117 0.0548185841681058
118 0.0548187415887018
119 0.0548210599124016
120 0.0548251868584069
121 0.0548292190596896
122 0.0548324898184839
123 0.0548337173757611
124 0.0548372253483423
125 0.0548372196258944
126 0.054837205319772
127 0.0548371852911945
128 0.0548377146154292
129 0.0548376030285831
130 0.0548375944449702
131 0.0548375944449702
132 0.0548375858613559
133 0.0548377060318338
134 0.0548376459466277
135 0.0548378576751562
136 0.0548378862870568
137 0.0548396487513517
138 0.0548455680423298
139 0.0548498190094048
140 0.0548552423760389
141 0.0548729618340575
142 0.0548819366324204
143 0.0548883487544218
144 0.0548954946863422
145 0.0548951688520884
146 0.0548944542965151
147 0.054897618278005
148 0.0548991759074165
149 0.0549058031979752
150 0.054908877943394
151 0.054909166549134
152 0.0549113696182131
153 0.0549125296942901
154 0.0549118439379945
155 0.0549172897498744
156 0.0549220493929079
157 0.0549323100772792
158 0.0549356261080342
159 0.0549399900536267
160 0.0549509983369727
161 0.0549585129722117
162 0.0549617360703917
163 0.0549647734319116
164 0.0549652615636437
165 0.0549654242732577
166 0.0549655812733059
167 0.0549672368828788
168 0.0549687896851593
169 0.0549698457940007
170 0.054970085556425
171 0.0549720607023189
172 0.0549761477730848
173 0.0549761905829722
174 0.0549760564452135
175 0.0549760564452135
176 0.0549760564452135
177 0.0549760564452135
178 0.0549760564452135
179 0.0549760564452135
180 0.0549760564452135
181 0.0549760564452135
182 0.0549760564452135
183 0.0549760564452135
184 0.0549760564452135
185 0.0549760564452135
186 0.0549760564452135
187 0.0549760564452135
188 0.0549760564452135
189 0.0549760564452135
190 0.0549760564452135
191 0.0549760564452135
192 0.0549760564452135
193 0.0549760564452135
194 0.0549760564452135
195 0.0549760564452135
196 0.0549760564452135
197 0.0549760564452135
198 0.0549760564452135
199 0.0549760564452135
200 0.0549760564452135
201 0.0549760564452135
202 0.0549760564452135
203 0.0549760564452135
204 0.0549760564452135
205 0.0549760564452135
206 0.0549760564452135
207 0.0549760564452135
208 0.0549760564452135
209 0.0549760564452135
210 0.0549760564452135
211 0.0549760564452135
212 0.0549760564452135
213 0.0549760564452135
214 0.0549760564452135
215 0.0549760564452135
216 0.0549760564452135
217 0.0549760564452135
218 0.0549760564452135
219 0.0549760564452135
220 0.0549760564452135
221 0.0549760564452135
222 0.0549760564452135
223 0.0549760564452135
224 0.0549760564452135
225 0.0549760564452135
226 0.0549760564452135
227 0.0549760564452135
228 0.0549760564452135
229 0.0549760564452135
230 0.0549760564452135
231 0.0549760564452135
232 0.0549760564452135
233 0.0549760564452135
234 0.0549760564452135
235 0.0549760564452135
236 0.0549760564452135
237 0.0549760564452135
238 0.0549760564452135
239 0.0549760564452135
240 0.0549760564452135
241 0.0549760564452135
242 0.0549760564452135
243 0.0549760564452135
244 0.0549760564452135
245 0.0549760564452135
246 0.0549760564452135
247 0.0549760564452135
248 0.0549760564452135
249 0.0549760564452135
250 0.0549760564452135
251 0.0549760564452135
252 0.0549760564452135
253 0.0549760564452135
254 0.0549760564452135
255 0.0549760564452135
256 0.0549760564452135
257 0.0549760564452135
258 0.0549760564452135
259 0.0549760564452135
260 0.0549760564452135
261 0.0549760564452135
262 0.0549760564452135
263 0.0549760564452135
264 0.0549760564452135
265 0.0549760564452135
266 0.0549760564452135
267 0.0549760564452135
268 0.0549760564452135
269 0.0549760564452135
270 0.0549760564452135
271 0.0549760564452135
272 0.0549760564452135
273 0.0549760564452135
274 0.0549760564452135
275 0.0549760564452135
276 0.0549760564452135
277 0.0549760564452135
278 0.0549760564452135
279 0.0549760564452135
280 0.0549760564452135
281 0.0549760564452135
282 0.0549760564452135
283 0.0549760564452135
284 0.0549760564452135
285 0.0549760564452135
286 0.0549760564452135
287 0.0549760564452135
288 0.0549760564452135
289 0.0549760564452135
290 0.0549760564452135
291 0.0549760564452135
292 0.0549760564452135
293 0.0549760564452135
294 0.0549760564452135
295 0.0549760564452135
296 0.0549760564452135
297 0.0549760564452135
298 0.0549760564452135
299 0.0549760564452135
300 0.0549760564452135
301 0.0549760564452135
302 0.0549760564452135
303 0.0549760564452135
304 0.0549760564452135
305 0.0549760564452135
306 0.0549760564452135
307 0.0549760564452135
308 0.0549760564452135
309 0.0549760564452135
310 0.0549760564452135
311 0.0549760564452135
312 0.0549760564452135
313 0.0549760564452135
314 0.0549760564452135
315 0.0549760564452135
316 0.0549760564452135
317 0.0549760564452135
318 0.0549760564452135
319 0.0549760564452135
320 0.0549760564452135
321 0.0549760564452135
322 0.0549760564452135
323 0.0549760564452135
324 0.0549760564452135
325 0.0549760564452135
326 0.0549760564452135
327 0.0549760564452135
328 0.0549760564452135
329 0.0549760564452135
330 0.0549760564452135
331 0.0549760564452135
332 0.0549760564452135
333 0.0549760564452135
334 0.0549760564452135
335 0.0549760564452135
336 0.0549760564452135
337 0.0549760564452135
338 0.0549760564452135
339 0.0549760564452135
340 0.0549760564452135
341 0.0549760564452135
342 0.0549760564452135
343 0.0549760564452135
344 0.0549760564452135
345 0.0549760564452135
346 0.0549760564452135
347 0.0549760564452135
348 0.0549760564452135
349 0.0549760564452135
350 0.0549760564452135
351 0.0549760564452135
352 0.0549760564452135
353 0.0549760564452135
354 0.0549760564452135
355 0.0549760564452135
356 0.0549760564452135
357 0.0549760564452135
358 0.0549760564452135
359 0.0549760564452135
360 0.0549760564452135
361 0.0549760564452135
362 0.0549760564452135
363 0.0549760564452135
364 0.0549760564452135
365 0.0549760564452135
366 0.0549760564452135
367 0.0549760564452135
368 0.0549760564452135
369 0.0549760564452135
370 0.0549760564452135
371 0.0549760564452135
372 0.0549760564452135
373 0.0549760564452135
374 0.0549760564452135
375 0.0549760564452135
376 0.0549760564452135
377 0.0549760564452135
378 0.0549760564452135
379 0.0549760564452135
380 0.0549760564452135
381 0.0549760564452135
382 0.0549760564452135
383 0.0549760564452135
384 0.0549760564452135
385 0.0549760564452135
386 0.0549760564452135
387 0.0549760564452135
388 0.0549760564452135
389 0.0549760564452135
390 0.0549760564452135
391 0.0549760564452135
392 0.0549760564452135
393 0.0549760564452135
394 0.0549760564452135
395 0.0549760564452135
396 0.0549760564452135
397 0.0549760564452135
398 0.0549760564452135
399 0.0549760564452135
400 0.0549760564452135
401 0.0549760564452135
402 0.0549760564452135
403 0.0549760564452135
404 0.0549760564452135
405 0.0549760564452135
406 0.0549760564452135
407 0.0549760564452135
408 0.0549760564452135
409 0.0549760564452135
410 0.0549760564452135
411 0.0549760564452135
412 0.0549760564452135
413 0.0549760564452135
414 0.0549760564452135
415 0.0549760564452135
416 0.0549760564452135
417 0.0549760564452135
418 0.0549760564452135
419 0.0549760564452135
420 0.0549760564452135
421 0.0549760564452135
422 0.0549760564452135
423 0.0549760564452135
424 0.0549760564452135
425 0.0549760564452135
426 0.0549760564452135
427 0.0549760564452135
428 0.0549760564452135
429 0.0549760564452135
430 0.0549760564452135
431 0.0549760564452135
432 0.0549760564452135
433 0.0549760564452135
434 0.0549760564452135
435 0.0549760564452135
436 0.0549760564452135
437 0.0549760564452135
438 0.0549760564452135
439 0.0549760564452135
440 0.0549760564452135
441 0.0549760564452135
442 0.0549760564452135
443 0.0549760564452135
444 0.0549760564452135
445 0.0549760564452135
446 0.0549760564452135
447 0.0549760564452135
448 0.0549760564452135
449 0.0549760564452135
450 0.0549760564452135
451 0.0549760564452135
452 0.0549760564452135
453 0.0549760564452135
454 0.0549760564452135
455 0.0549760564452135
456 0.0549760564452135
457 0.0549760564452135
458 0.0549760564452135
459 0.0549760564452135
460 0.0549760564452135
461 0.0549760564452135
462 0.0549760564452135
463 0.0549760564452135
464 0.0549760564452135
465 0.0549760564452135
466 0.0549760564452135
467 0.0549760564452135
468 0.0549760564452135
469 0.0549760564452135
470 0.0549760564452135
471 0.0549760564452135
472 0.0549760564452135
473 0.0549760564452135
474 0.0549760564452135
475 0.0549760564452135
476 0.0549760564452135
477 0.0549760564452135
478 0.0549760564452135
479 0.0549760564452135
480 0.0549760564452135
481 0.0549760564452135
482 0.0549760564452135
483 0.0549760564452135
484 0.0549760564452135
485 0.0549760564452135
486 0.0549760564452135
487 0.0549760564452135
488 0.0549760564452135
489 0.0549760564452135
490 0.0549760564452135
491 0.0549760564452135
492 0.0549760564452135
493 0.0549760564452135
494 0.0549760564452135
495 0.0549760564452135
496 0.0549760564452135
497 0.0549760564452135
498 0.0549760564452135
499 0.0549760564452135
500 0.0549760564452135
};
\addlegendentry{Cones};

\end{axis}
\end{tikzpicture}%